\documentclass{article}

 \usepackage[preprint]{neurips_2024}

\usepackage{microtype}
\usepackage{graphicx}
\usepackage{subfigure}
\usepackage{booktabs} 

\usepackage[utf8]{inputenc} 
\usepackage[T1]{fontenc}    
\usepackage{url}            
\usepackage{booktabs}       
\usepackage{amsfonts}       
\usepackage{nicefrac}       
\usepackage{microtype}      
\usepackage{xcolor}         

\usepackage[nodisplayskipstretch]{setspace}

\usepackage{hyperref}

\usepackage{amsmath}
\usepackage{amssymb}
\usepackage{mathtools}
\usepackage{amsthm}

\usepackage[capitalize,noabbrev]{cleveref}

\theoremstyle{plain}
\newtheorem{theorem}{Theorem}[section]

\theoremstyle{definition}

\theoremstyle{remark}

\newcommand{\eins}{\leavevmode\hbox{\small1\kern-3.8pt\normalsize1}}

\usepackage[textsize=tiny]{todonotes}

\def\be{\begin{equation}}
\def\ee{\end{equation}}
\def\bes{\begin{subequations}}
\def\ees{\end{subequations}}
\def\bea{\begin{eqnarray}}
\def\eea{\end{eqnarray}}
\def\bry{\begin{array}}
\def\ery{\end{array}}
\def\bit{\begin{itemize}}
\def\eit{\end{itemize}}
\def\ben{\begin{enumerate}}
\def\een{\end{enumerate}}
\def\nn{\nonumber}

\def\tr{\textrm{Tr}}

\def\({\left(}
\def\){\right)}

\def\sigA{\Sigma_A}
\def\sigB{\Sigma_B}

\newcommand{\norm}[1]{\| #1 \|}

\crefformat{equation}{Eq.~(#2#1#3)}
\Crefformat{equation}{Eq.~(#2#1#3)}

\crefformat{section}{Sec.~#2#1#3}
\Crefformat{Section}{Sec.~#2#1#3}
\crefformat{figure}{Fig.~#2#1#3}
\Crefformat{Figure}{Fig.~#2#1#3}

\Crefformat{appendix}{App.~#2#1#3}
\Crefformat{Appendix}{App.~#2#1#3}
\Crefformat{app}{App.~#2#1#3}

\def\be{\begin{equation}}
\def\ee{\end{equation}}
\def\bes{\begin{subequations}}
\def\ees{\end{subequations}}
\def\bea{\begin{eqnarray}}
\def\eea{\end{eqnarray}}
\def\bry{\begin{array}}
\def\ery{\end{array}}
\def\bit{\begin{itemize}}
\def\eit{\end{itemize}}
\def\ben{\begin{enumerate}}
\def\een{\end{enumerate}}
\def\nn{\nonumber}

\def\tr{\textrm{Tr}}

\def\({\left(}
\def\){\right)}

\newcommand{\thmref}[1]{Theorem~\cref{#1}}

\renewcommand{\eqref}[1]{Eq.~(\cref{#1})}

\newcommand{\stam}[1]{}

\newcommand{\bx}{\mathbf{x}}

\newcommand{\btheta}{{\boldsymbol{\theta}}}

\newcommand{\cl}{{\cal L}}

\newcommand{\reals}{{\mathbb R}}

\title{Classifying Overlapping Gaussian Mixtures in High Dimensions: From Optimal Classifiers to Neural Nets}

\author{%
  Khen Cohen$^*$,~Yaron Oz\\
  Raymond and Beverly Sackler School of Physics and Astronomy\\
  Tel-Aviv University\\
  Tel-Aviv 69978, Israel \\
  \texttt{khencohen@mail.tau.ac.il} \\
  \And
  Noam Levi\thanks{Equal Contribution}\\
  \'Ecole Polytechnique F\'ed\'erale de Lausanne~(EPFL)\\
  Switzerland\\
  \texttt{noam.levi@epfl.ch} \\
}

\begin{document}

\maketitle

%

\begin{abstract}

We derive closed-form expressions for the Bayes optimal decision boundaries in binary classification of high dimensional overlapping Gaussian mixture model (GMM) data, and show how they depend on the eigenstructure of the class covariances, for particularly interesting structured data.
We empirically demonstrate, through experiments on synthetic GMMs inspired by real-world data, that deep neural networks trained for classification, learn predictors which approximate the derived optimal classifiers. 
We further extend our study to networks trained on authentic data, observing that decision thresholds correlate with the covariance eigenvectors rather than the eigenvalues, mirroring our GMM analysis. This provides theoretical insights regarding neural networks' ability to perform probabilistic inference and distill statistical patterns from intricate distributions.

%
%

%


\end{abstract}

\section{Introduction}
\label{sec:intro}

There is a well-accepted understanding in machine learning that inherent correlations or statistical structure within the data play a crucial role in enabling effective modeling. The presence of structure in the data provides important context that machine learning algorithms can leverage for accurate and meaningful knowledge extraction and generalization. Harnessing any inherent structure is widely seen as an important factor for achieving successful learning outcomes.

Determining the characteristics of natural datasets that contribute to effective training and good generalization has tremendous theoretical and practical significance. 
Regrettably, real-world datasets are typically viewed as samples drawn from some unknown, intricate, high-dimensional underlying population, rendering analyses that depend on accurately modeling the true distribution enormously challenging. 
Connecting properties of the data to learning outcomes is difficult because real data does not readily yield its generative process or population distribution. 

While fully characterizing natural data distributions remains difficult, progress can still be made by considering simplified models that capture key aspects of the true distribution. 
One approach is to model the data via the moment expansion of the underlying population. For a random vector $X \in \mathbb{R}^d$ representing a single sample, its distribution can be described by the moment generating function $M_X(t)=\sum_{n=0}^\infty t^n \mathbb{E} (X^n)/ n!$. Approximating this by retaining a finite number of moments provides a tractable model of the data distribution.

A simple starting point is the Gaussian model, which posits that a dataset's statistical properties can be fully captured by its first two moments - the mean vector and sample covariance matrix. This assumes higher-order dependencies are negligible such that the distribution is characterized solely by its first and second central statistical moments.

In the last decade, analyzing the behavior of machine learning algorithms on idealized i.i.d. Gaussian datasets has emerged as an important area of research in high-dimensional statistics~\citep{donoho2009observed,KoradaMontanari2011, Monajemi2013, candes2020phase, Bartlett30063}. Studying these simplified synthetic distributions has provided valuable insights, such as neural network scaling laws~\citep{maloney2022solvable, kaplan2020scaling}, universal convergence properties~\citep{seddik2020random} and an improved understanding of the "Double Descent" phenomenon~\citep{montanari2022universality}. Examining learning on Gaussian approximations of real data distributions has helped establish foundational understandings of algorithmic behavior in high-dimensional settings.

A particular case of interest is that of data which is divided into a set number of classes. In certain instances, the data can be described by a mixture model, where each sample is generated separately for each class. The simplest example of such distributions is that of a Gaussian Mixture Model (GMM), on which we focus in this work.

Gaussian mixtures are a popular model in high-dimensional statistics since, besides being an universal approximator, they often lead to mathematically tractable problems. Indeed, a recent line of work has analyzed the asymptotic performance of a large class of machine learning problems in the proportional high-dimensional limit under the Gaussian mixture data assumption, see e.g. 
\cite{mai2019high,mignacco2020role,taheri2020optimality,kini2021phase,wang2021benign,refinetti2021classifying,loureiro2021learning_gm}.

In light of the insights gained by studying the {\it unsupervised}, 
perspective on GMM classification, we focus here on a complementary direction.
Namely, we consider the 
{\it supervised} setting, where a single neural network is trained on a
dataset generated from a GMM, to perform binary classification where the true labels are given. 
%
To connect the GMM with real-world data, we restrict ourselves to the case of strictly non-linearly separable GMMs, such that the class means difference is negligible compared to the class covariance difference. We refer to this setup as {\it overlapping GMM classification}.

Our {\bf main contributions} are as follows:
\setlength{\leftmargini}{16pt}
\vspace{-0.2cm}
\begin{itemize}\setlength\itemsep{-.1\itemsep}
    \item 
    In \cref{sec:GMM}, we derive the Bayes Optimal Classifiers (BOCs) and decision boundaries for overlapping GMM classification problems, both in the population and the empirical limits.
    \item
    In \cref{sec:data_modeling}, under the assumptions of correlated features and Haar distributed eigenvectors, we are able to provide approximate closed form equations for the decision boundaries and discriminator distributions, as a function of the eigenvalues and eigenvectors of the different class covariances.
    \item 
    In \cref{sec:NN_optimal},
    we present empirical evidence, demonstrating that some networks trained to perform binary classification on a GMM, approximate the BOC.  
    We further relate these observations to existing results, namely convergence of homogeneous networks to a KKT point.
    \item 
    In \cref{sec:Realistic_optimal}, we demonstrate empirically that for high dimensional data, the covariance eigenvectors and not the eigenvalues determine the classification threshold for deep neural networks trained on real-world datasets, and provide an explanation inspired by our results on GMMs.
    In \cref{conclusions} we summarize our conclusions and outlook.
\end{itemize}

\setlength{\belowdisplayskip}{4pt} 
\setlength{\belowdisplayshortskip}{4pt}
\setlength{\abovedisplayskip}{4pt} 
\setlength{\abovedisplayshortskip}{4pt}

\vspace{-.2cm}
\section{Background and Related Work}
\label{back}
\vspace{-.2cm}

There has been significant work aimed at understanding the classification capabilities of Gaussian mixture models (GMMs), that is, recovering the cluster label for each data point rather than using labels provided by a teacher. For binary classification problems, examples include methods proposed by \citet{mai.liao.ea_2019_large}, \citet{mignacco20}, and \citet{deng.kammoun.ea_2022_model}, with the latter also demonstrating an equivalence between classification and single-index models. In multi-class settings, \citep{thrampoulidis.oymak.ea_2020_theoretical} analyzed the performance of ridge regression classifiers. The most general results in this area are from Gaussian Mix Group \citep{gaussian_mix_group}, which considers the use of GMMs with any convex loss function for classification. Additional results regarding kernels and GMMs include~\citep{Couillet2018, Liao_2019, Kammoun2023,refinetti2021classifying}, precise asymptotics~\citep{NEURIPS2021_543e8374} and the connection between GMMs and real-world vision data~\citep{Ingrosso_2022}.
Additionally, work has been done to understand Quadratic Discriminant Analysis (QDA) in high dimensions~\citep{elkhalil2017asymptotic,ghojogh2019linear, Das_2021}, which we rely heavily upon in this work.

\vspace{-.2cm}
\section{Overlapping Gaussian Mixtures in High Dimensions}
\label{sec:GMM}
\vspace{-.2cm}

Consider the task of performing binary classification on samples drawn from a two class GMM, where the underlying Gaussian distributions have similar means but different covariance matrices.
We generate two datasets $\mathcal{D}_A, \mathcal{D}_B$ of equal size $N$, and the same data dimensions $x\in \mathbb{R}^d$. 
Here, each sample vector is drawn from a jointly normal distribution, either $x\sim \mathcal{N}(\mu_A,\sigA)$ or $x\sim \mathcal{N}({\mu}_B,\sigB)$, where the two means and covariance matrices distinguish between the two classes. 
In this setup, we assign a label value $y=1$ to samples drawn from class $A$ and $y=-1$ for the $B$ class samples. Throughout this work, we consider the large number of features and large number of samples limit $d, N \to \infty$, while the ratio $d/N = \gamma \in \mathbb{R^+}$ is constant. We denote $\gamma\to0$ as the {\it population} limit, while $\gamma \to 1$ as the {\it empirical} limit.

In the following sections, we discuss the optimal classifiers obtained first from the population and then the empirical data distributions. We then apply our results to a simplified model dataset which captures some of the properties of real-world datasets, and distinguish the roles of the covariance eigenvectors and eigenvalues for this case.

\begin{figure*}
    \centering
    \includegraphics[width=.32\linewidth]{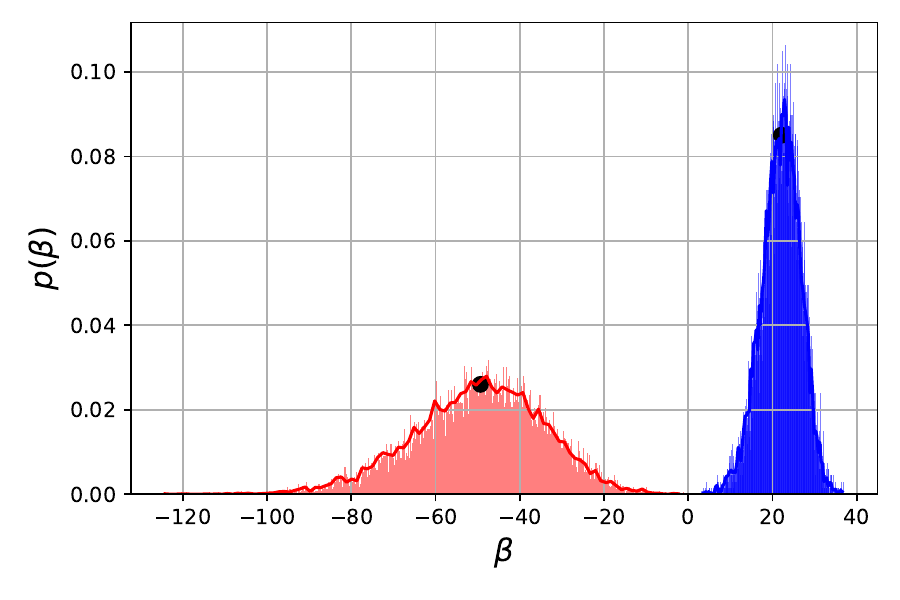}~
    \includegraphics[width=.32\linewidth]{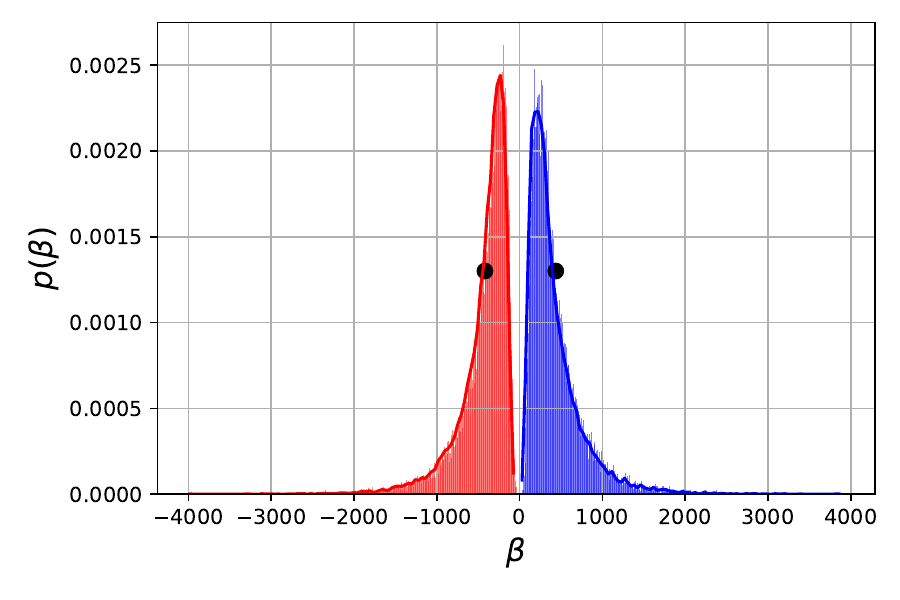}~
    \includegraphics[width=.32\linewidth]{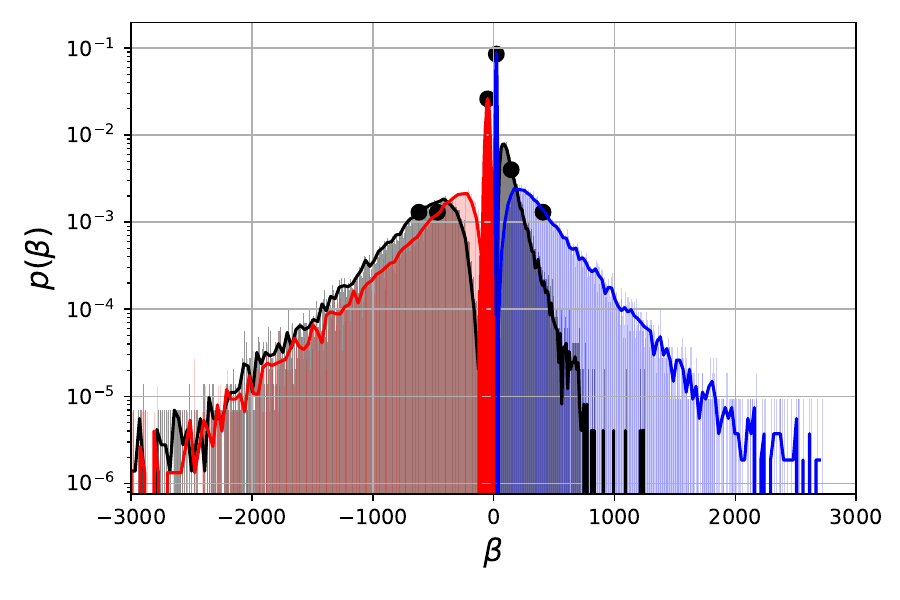}
    \vspace{-.4cm}
    \caption{Probability density function of $\beta(x)$, evaluated on GMMs with different class covariance matrices. {\bf Blue} and {\bf red} bins indicate samples drawn from the classes $A$ and $B$, respectively. Solid curves represent the numerical evaluation of \cref{eq:beta_def} as a generalized $\chi^2$ distribution. Dots indicate the values given by \cref{eq:mean_classes_general}, as well as by~\cref{eq:BOC_spectral_diff}. 
    {\bf Left:} $\beta$ distribution for covariances with the same basis but different spectra. {\bf Center:} $\beta$ distribution for covariances with the same spectrum but different random bases. 
    {\bf Right:} $\beta$ distribution comparison for covariances with both different spectra and different bases, shown in {\bf gray}. 
    Here, we take $\alpha_A = 0.5, \Delta \alpha = -0.3, d=100$.
    }
    \label{fig:beta_dist}
    \vspace{-0.4cm}
\end{figure*}

\vspace{-.2cm}
\subsection{Optimal Classification on Population Data}
\label{sec:BOC}
\vspace{-.2cm}

The Bayes-optimal classifier (BOC) assigns each sample
to the class that maximizes the expected value gain or its logarithm defined as
\begin{align}
    G(X=x)
    =
    \arg \max_{Y=y}p(X=x| Y=y)p(Y=y)
    =
    \arg \max_{Y=y} \log( p_{X|Y})=
    \arg \max_{Y=y} \beta_Y
    ,
\end{align}
where $p(Y=y)=1/C$ is the class density with $C$ the number of classes in the case of a balanced dataset. Here, $p(X=x| Y=y)=\exp{(\beta_Y)}$ is the conditional in-class density.
The decision boundary between any two classes $y,y'\in C$ is given when ${\beta_{Y=y}}(X=x)={\beta_{Y=y'}}(X=x)$, where the probabilities of a sample being classified as $y$ or $y'$ are equal. For a two-class GMM, this is obtained by simply equating the class conditional probability densities, i.e., $p_A=p_B$,
leading to the {\it quadratic} decision rule, derived in~\cite{elkhalil2017asymptotic} as well as~\cite{Das_2021}, to pick class $A$ if
\begin{align}
\label{eq:BOC}
&\beta(x) =
\frac{1}{2}
(
x^T Q x  -2 q^T x + c
)>0,
\quad
Q = 
(\sigB^{-1} - \sigA^{-1}), \quad
q = 
( \sigB^{-1}\mu_B - \sigA^{-1} \mu_A ) ,
\\
\nn
&c = 
\mu_B^T \sigB^{-1}\mu_B
-\mu_A^T \sigA^{-1}\mu_A
-
\log \left(\frac{|{\sigA }|}{|\sigB| }\right) \ ,
\end{align}
where $|O|$ is the determinant of the matrix $O$.
This quadratic $\beta(x)$ is the Bayes classifier, or the Bayes decision variable that, when compared to zero, maximizes expected gain.

Note that the $q^T x$ term in~\cref{eq:BOC} determines the {\it linear} decision boundary, while $x^T Q x$ determines the {\it nonlinear} part. 
For the rest of this work, we consider the {\bf nonlinear, overlapping mixture configuration, where $\mu_A=\mu_B=\mu$, while $\Sigma_A\neq \Sigma_B$}, as a proxy for the behavior of many real world datasets~\citep{Schilling_2021}.
Without loss of generality, we may set $\mu=0$, since we can always shift away the mean of the distributions, resulting in the simpler form of $\beta(x)$ 
\begin{align}
\label{eq:beta_def}
   \beta(x)
  =
  \frac{1}{2}
  (
  x^T Q x + 
    c
    )
    =
      \frac{1}{2}
    (
   \tr (Q x x^T ) 
   +
   c
   )
   =
   \frac{1}{2}
    \sum_i
    q_i  \tilde{\chi}^2_{1}
    -
    \frac{a}{2}
    ,
\end{align}
where the last transition in~\cref{eq:beta_def} is valid since $\beta$ is given by a sum of $\chi^2$ distributed random variables, known as the generalized chi-squared distribution~\citep{Das_2021},
where $a\sim \mathcal{N}(\log(|\sigB^{-1}\sigA|),0)$, $q_i$ are the eigenvalues of $\Sigma_Y^{1/2} Q\Sigma_Y^{1/2}$, and $Y=A,B$. The probability density function (PDF) of $\beta$ does not have a closed analytic form, but can be obtained numerically by various methods~\citep{ferrari2019note,Das_2021}. 
In \cref{fig:beta_dist}, we show the $\beta$-distribution for high-dimensional Gaussian data with different covariances, obtained both empirically from \cref{eq:beta_def} and by drawing from a sum of $\chi^2$ variables matching \cref{eq:beta_def}.

\setlength{\belowdisplayskip}{4pt} 
\setlength{\belowdisplayshortskip}{4pt}
\setlength{\abovedisplayskip}{4pt} 
\setlength{\abovedisplayshortskip}{4pt}

While the full PDF of $\beta$ cannot be written in a simple form, the empirical expectation value on a given set of measurements can be derived. 
Namely, given a set of $N$ samples $x_i, i=1,\ldots,N$, taken from a normal distribution $\mathcal{N}(0,M), M\in\mathbb{R}^{d\times d}$, the expectation value of the Bayes classifier $\beta$ on these samples is simply 
\begin{align}
    \langle \beta_M \rangle
    =
    \frac{1}{2}
    \left[ \tr   ( QM )
    -
    \log \left(
    {|\sigB^{-1}|
    {|{\sigA }|} }\right)
    \right] ,
\end{align}
where $\langle \cdot \rangle $ indicates averaging over the data distribution.
It can be used to define what it means for a sample to belong to class $A$ or $B$, by computing the average distance between the expectation value of $\beta$ on any dataset and the class values, given by
\begin{align}
\label{eq:mean_classes_general}
\langle \beta_A  \rangle
    =
    -d + \tr ( \sigB^{-1} \sigA )
    +c
    > 0, 
    \quad
    \langle \beta_B  \rangle
    =
    d - \tr ( \sigA^{-1} \sigB) 
    +
    c
    < 0,
\end{align}
where we averaged over $N\to \infty$ samples from the $A,B$ distributions, and $c=-\log(|\sigB^{-1}\sigA|)$.
We show these values as black dots in \cref{fig:beta_dist}.

\begin{figure*}[t]
    \centering
    \includegraphics[width=.99\linewidth]{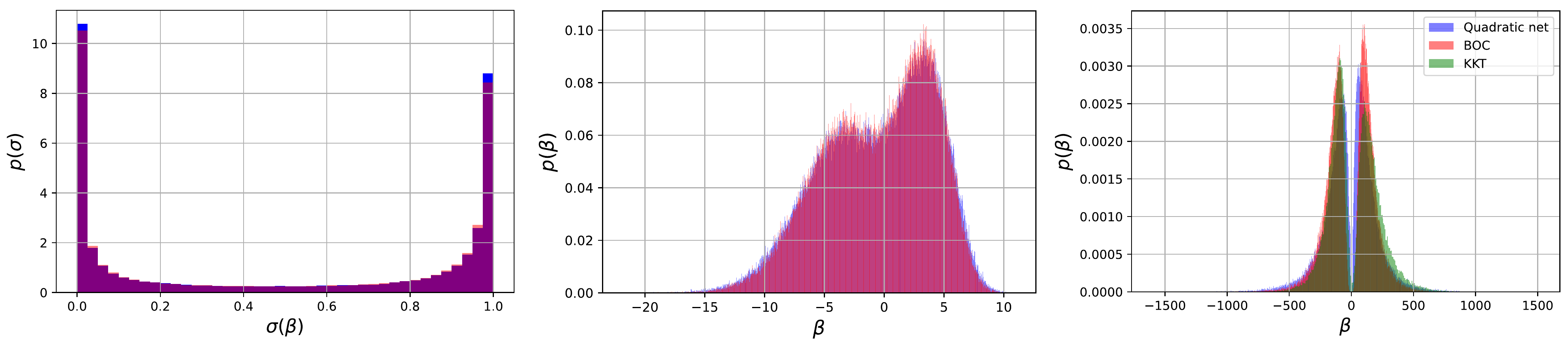}~
        \vspace{-.4cm}
    \caption{Comparison between BOC and quadratic network trained on data with different covariance spectrum. {\bf Left:} accuracy, measured as the sigmoid function acting on the network/BOC output. {\bf Center:} network/BOC output distribution.
    {\bf Blue} indicates network results while {\bf red} is the BOC prediction. Here, the network is trained with $d_h=100$ and $N=200k$ to $91.27\%$ training accuracy and reaches $91.1\%$ with $\alpha_A=0.2$ and $\alpha_B=0.3$.
    {\bf Right:} Results for KKT convergence of a quadratic NN trained on an $\alpha=0.2$ correlated GMM, where the class covariances share the same spectrum but are given a different random basis.
    {\bf Red} indicates the BOC prediction, {\bf blue} is the quadratic network output before the softmax function, when trained with $d_h=100$, and {\bf green} are the KKT predictions with $\lambda_a
    \propto 1/N$ and $d_h=100$. The data dimensions are similar to the main text, i.e., $d=100$, using $N=100k$ samples, the network reached $100\%$ training and $99.9\%$ test accuracy.
    } 
    \label{fig:network_optimal}
        \vspace{-0.5cm}
\end{figure*}


\vspace{-.2cm}
\subsection{Empirical Optimal Classification}
\label{sub:empirical_covariances}
\vspace{-.2cm}

So far, we have assumed that we have access to the population covariances $\sigA,\sigB$. Here, we comment on the classification problem changes when relaxing this assumption and consider the empirical covariances which are obtained at finite dimension to sample number ratio $\gamma = const.$

In particular, we consider the case where we have access only to finitely sampled versions of the population moments, but have infinitely many samples of this noisy distribution. 
Here, the distributions in question for classes $A,B$ are still gaussian, but their respective moments are not the population ones, implying that the BOC is deformed.
The covariance matrices we have access to are given by measurements, that define the empirical covariances for each class $\Sigma_{N,Y} = \frac{1}{N} X_YX_Y^T\simeq \Sigma_Y+\sqrt{\gamma} R_Y$, where $X_Y\in\mathbb{R}^{d\times N}$ is the design matrix, and $R_Y$ is a random matrix with $\mathcal{O}(1)$ eigenvalues~\citep{Biroli_2023}\footnote{ This result holds in the limit of $d^2 \ll N$. In the case of $d\ll N$, $Q_N$ is a Wishart matrix, which requires more careful treatment such as given in~\cite{Liao_2019,Tiomoko2019,zhangWang}, which we leave for future work.}.
We can therefore rewrite the empirical BOC as
\begin{align}
\label{eq:BOC_empirical}
    \!\!\! \beta_N(x)
    &\simeq
    \frac
    {x^T  \left(
    Q_N
    \right) x}
    {2}
    +
    \frac{c_N}{2}
    ,
    ~
    Q_N
    \simeq
    Q
    +\sqrt{\gamma }
    ({R_B}
    - {R_A}),
    ~
    c_N \!
    =
    c
    -
    \log \left(
    \frac{|\mathbb{I}+\sqrt{\gamma }{\sigB R_B}|}{|\mathbb{I}
    +\sqrt{\gamma }{\sigA  R_A}| }\right),
\end{align}
where we expand to leading order in $\gamma=\frac{d}{N}$. 
This implies that the empirical $\beta_N$ distribution remains a generalized $\chi^2$ with the substitution of coefficients to be the eigenvalues of $\Sigma_Y^{1/2 }Q_N\Sigma_Y^{1/2 }$ and the constant $c$ is shifted by the $\log$ term.
Therefore, the empirical deviation from the BOC
$\Delta \beta_N = \|\beta_N- \beta \|$ is estimated by
\begin{align}
\!\!\!\!
    \Delta \beta_N
    &\simeq
    \sqrt{\gamma }
    |  
    \tr({R_B} (xx^T- \sigB )
    - {R_A}(xx^T-\sigA))|
    ,
\end{align}
which decreases with $\sqrt{\gamma}$.
This setting is exhibited in~\cref{fig:beta_dist}, which is meant to mimic the real-world setting, where we do not have access to the population covariance, and we can only estimate how close the empirical moments are to the population ones.

\vspace{-.2cm}
\section{Analysis for a Toy Model of Complex Data}
\label{sec:data_modeling}
\vspace{-.2cm}

In order to connect the BOC with real world classification tasks, we focus on a specific modeling of natural data.
Here, we consider a simple model of a correlated GMM inspired by the neural scaling law literature~\citep{maloney2022solvable, montanari2022universality,levi2023underlying} as well as the signal recovery literature~\citep{loureiro2021learning,Loureiro_2022}. Concretely, we assume a power law scaling spectrum, 
and a different basis matrix for each class
\begin{align}
    \Lambda_{Y,ij} = i^{-1-\alpha_Y}\delta_{ij},
    \qquad
    \Sigma_Y = O_Y^T \Lambda O_Y ,
    \label{CM}
\end{align}
where $\delta_{ij}$ is the $d$ dimensional identity matrix and $O_Y$ is a random orthogonal matrix.
In the following sections, 
we distinguish the role of eigenvalues and eigenvectors in determining the BOC for overlapping GMMs.

\vspace{-.2cm}
\subsection{Diagonal Correlated Covariances}
\vspace{-.2cm}

As a first example, we consider the case where $\sigA, \sigB$ are both diagonal matrices with different eigenvalues, such that $O_A=O_B=\mathbb{I}$, and $\sigA=\Lambda_A, \sigB=\Lambda_B$.
Given a model of power law scaling dataset $A$ with $\alpha_A$, we can define the power law exponent of the dataset $B$ to be $\alpha_B =\alpha_A +\Delta \alpha$. Here, we can obtain explicit expressions for $\beta$ as
\begin{align}
\label{eq:BOC_spectral_diff}
   \!\!\!
   \langle \beta_A \rangle
    \!=\!
   \frac{1}{2} 
   \left(
   H_d^{(-\Delta \alpha )}
   \!\!
   -
   \!d
   -\Delta \alpha  \log (\Gamma (d+1))
   \right),
   ~
       \langle \beta_B \rangle
    \!=\!
    \frac{1}{2} 
   \left(
    d
    -H_d^{(\Delta \alpha )}
    \!\!
    -\Delta \alpha  \log \left(\Gamma(d+1)\right)
    \right), 
\end{align}
where $H_d^{i}$ is the Harmonic number, and $\Gamma(x)$ is the Gamma function. 
In the limit of $d\to\infty$, one obtains that $\langle \beta_A \rangle \propto \frac{1}{4} \left(2 d \left(\frac{d^{\Delta \alpha }}{\Delta \alpha +1}-1\right)-(2 d \Delta \alpha +\Delta \alpha ) \log (d)\right)$ at leading order. This is a rapidly growing function in $d$, showing that even a small difference in spectra can be magnified simply by dimensionality, leading to correct classification.

\vspace{-.2cm}
\subsection{Rotated Correlated Covariances}
\vspace{-.2cm}

Next, we consider the case where $\sigA, \sigB$ share the same spectrum, but are rotated with respect to one another, such that $\sigA = O_A^T \Lambda O_A, \sigB= O_B^T \Lambda O_B$. The sample 
expectation for the BOC reads
\begin{align}
   \langle \beta_A \rangle
   =
   -  \langle \beta_B \rangle
    = 
    \frac{1}{2}
    \left(
     \tr ( \mathcal{O}^T\Lambda^{-1} \mathcal{O} \Lambda )-d 
    \right) 
    ,
\end{align}
where $\mathcal{O} = O_A O_B^T$ is itself a random orthogonal matrix. In the limit of $d\to\infty$, we may replace the expectation value over the dataset with its integral with respect to the Haar measure on the rotation group, and obtain a closed form expression as
\begin{align}
\label{eq:rotation_classifier}
       \langle \beta_A \rangle_\mathcal{O}
    &=
    \frac{1}{2}
    \left[
    \frac{1}{d}\tr (\Lambda^{-1} )\tr (\Lambda )
    -d 
    \right]
     =
    \frac{1}{2} \left(
    \frac{H_d^{(-1-\alpha)} H_d^{(1+\alpha)}}{d}
    -d\right)
    ,
\end{align}
where the final expression is given for a power law scaling dataset.
In the limit of $d\to \infty$, \cref{eq:rotation_classifier} grows as $\langle \beta_A \rangle_\mathcal{O}\propto \frac{1}{2} \left(\frac{1}{\alpha  (\alpha +2)}+1\right) d-\frac{d^{\alpha +1} \zeta (\alpha +1)}{2 (\alpha +2)}$, which is a faster growing function than~\cref{eq:BOC_spectral_diff} for $\Delta \alpha/\alpha<1$ indicating that it requires a large difference in the relative eigenvalue scaling exponents $\Delta \alpha/ \alpha$ to shift the BOC distribution, when compared to the effect of a basis difference. 
We show this effect explicitly in the different panels of~\cref{fig:beta_dist}.
This implies that the eigenvectors are likely to play a more significant role than eigenvalues in datasets which are well modeled by the above setup.

\begin{figure*}[t!]
    \centering
    \includegraphics[width=.45\linewidth]{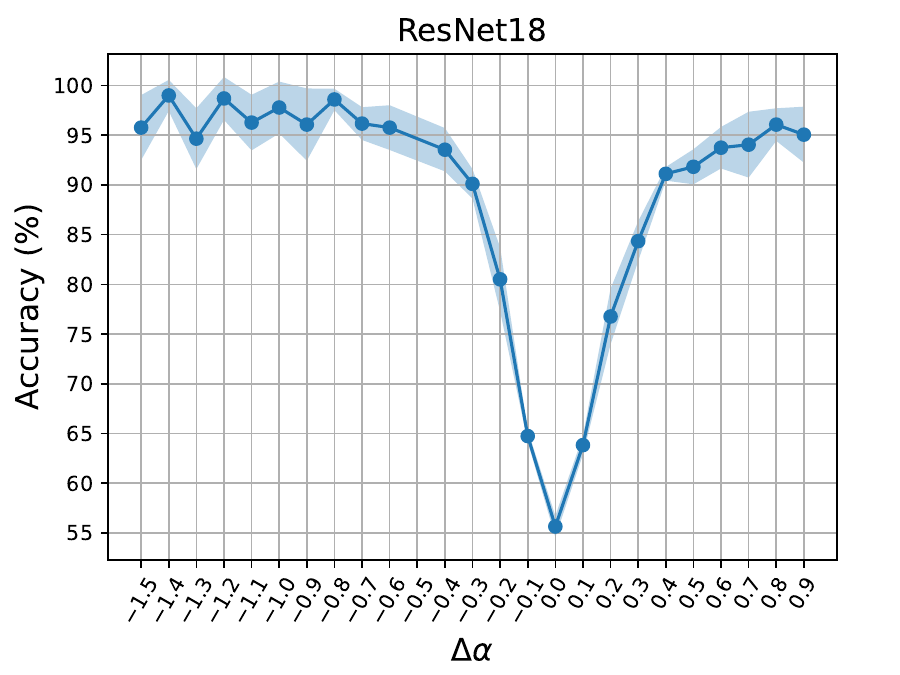}
    \includegraphics[width=.45\linewidth]{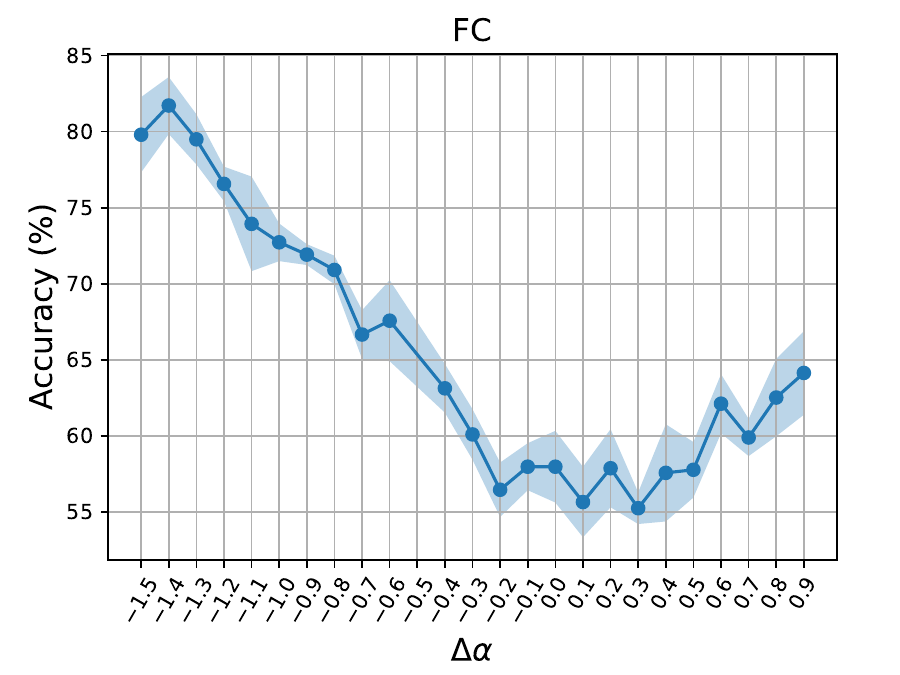}
        \vspace{-.3cm}
    \caption{Two Gaussian classification task, with the same eigenvectors but different eigenvalues bulk:
    the first one with $\alpha=0.5$ and the second with $\alpha+\Delta \alpha$. The model is FC, averaged over 5 training runs - the solid line represents the mean accuracy and shaded area represents 1-sigma error bars.}
    \label{fig:model_FC_alpha_dalpha}
\end{figure*}

\vspace{-.2cm}
\section{Neural Networks as Nearly Optimal Classifiers}
\label{sec:NN_optimal}
\vspace{-.2cm}

Having established the BOC behavior on overlapping GMMs, we can now ask how these conclusions translate to the learning process in some neural network architectures.
Namely, NNs have been shown to achieve optimal classification in some cases~\citep{Radhakrishnan_2023}. We demonstrate that NNs quite generically approximate the quadratic BOC, and provide some intuition as to how it may occur by appealing to the directional convergence of gradient descent to a KKT point.

\vspace{-.2cm}
\subsection{Neural Network Classifier}
\vspace{-.2cm}

Let $\mathcal{D} = \{(\bx_a,y_a)\}_{a=1}^N \subseteq \reals^d \times \{-1,1\}$ be a binary classification training dataset. Let $\Phi(\btheta;\cdot):\reals^d \to \reals$ be a neural network parameterized by $\btheta \in \reals^p$. 
For a loss function $\ell:\reals \to \reals$ the empirical loss of $\Phi(\btheta; \cdot)$ on the dataset $\mathcal{D}$ is 
$\cl(\btheta) := \sum_{a=1}^N \ell(y_a \Phi(\btheta; \bx_a))$.
We focus on 
the \emph{logistic loss} (a.k.a. \emph{binary cross entropy}), namely, $\ell(q) = \log(1+e^{-q})$. 

The network function $\Phi(\btheta;\cdot)$, can therefore be interpreted as the equivalent of the BOC variable $\beta$ in the previous sections. With this intuition in mind, we expect that neural networks trained to perform binary classification should converge to the BOC, i.e., $\Phi(\btheta;\cdot) \to \beta$ with sufficient with sufficient network expressivity, number of samples and successful optimization dynamics.

The simplest network architecture that one may employ is a linear classifier, where $\Phi(z)=z$. Such linear networks of any depth cannot approximate the BOC given by~\cref{eq:beta_def}, as it is fully nonlinear.

Next, we consider a non-trivial case, by taking a two layer network, with the activation function $\sigma(z)=z^2$, i.e. the quadratic activation.
Quadratic activations are natural in the context of signal processing, where often detectors can only measure the amplitude (and not phase) of a signal, e.g. the phase retrieval problem~\citep{jaganathan2015phase,Dong2023}. It has also gained in popularity recently as a prototypical non-convex optimization problem with strict saddles, e.g.~\cite{candes2014,Chen_2019,arnaboldi2024escaping,martin2023impact}.
It has also been shown such two-layer blocks can be used to simulate higher-order polynomial neural networks and sigmoidal activated neural networks~\citep{livni2014computational, pmlr-v84-soltani18a}.
Concretely, a sufficiently expressive network function to fully realize the BOC is
\setlength{\belowdisplayskip}{2pt} 
\setlength{\belowdisplayshortskip}{2pt}
\setlength{\abovedisplayskip}{6pt} 
\setlength{\abovedisplayshortskip}{6pt}
\begin{align}
    \Phi = v^T (W x)^2+ b
    =
    \sum_i\sum_{\mu,\nu}
    v_i  W_{i\mu }W_{i\nu} x_\mu x_\nu + b,
\end{align}
where $W\in\mathbb{R}^{d_h \times d}$, $v^T\in\mathbb{R}^{d_h}$ are weight matrices at the input and hidden layers, $d_h$ is the number of hidden units, and $b$ is the bias at the last layer. 
In this setup, a predictor function can be easily matched to the BOC by matching
\vskip -0.2in
\begin{align}
    \sum_{i=1}^{d}
    v_i W_{i\mu}W_{i\nu} 
    =   \frac{1}{2} Q_{\mu \nu} ,
    \qquad
    b = \frac{c}{2},
\end{align}
\vskip -0.1in
where the number of hidden neurons correspond to the minimal number required to match the rank of the BOC matrix $Q$, hence $d_h \geq d$.
Since this network is expressive enough to fully reproduce the BOC, our analysis in~\cref{sec:BOC} holds, and the importance of eigenvalues and eigenvectors is manifest.
We show the result of a quadratic network converging to the BOC in \cref{fig:network_optimal} 
where 
$d=d_h=100$.

\vspace{-.2cm}
\subsection{Karush–Kuhn–Tucker (KKT) Convergence}
\label{sec:kkt_convergence_net}
\vspace{-.2cm}

Another source of intuition for the value of $d_h$ may be found in the KKT convergence equation~\citep{ji2020directional,lyu2020gradient}, which states, in a nutshell, that the weights of a homogeneous neural network
\footnote{Homogeneity, i.e. $f(L \theta) = L^a f(\theta)$, where $f$ is the output of the network, $\theta$ are the network parameters, and $L,a \in \mathbb{R}$, a quadratic network satisfies this requirement as it is composed of monomials.},
trained with the logistic loss for binary classification, using an infinitesimal learning rate $\eta \to 0$ and infinite iterations $t\to \infty$, will converge in direction to a KKT fixed point given by
\begin{align}
\label{eq:kkt_conv}
    \tilde{\btheta} = \sum_{a=1}^N \lambda_a y_a \nabla_\btheta  \Phi(\tilde{\btheta}; \bx_a),
\end{align}
where $\tilde{\btheta}$ are the network weights at convergence, $\bx_a$ are the samples and $y_a$ are the sample labels. Here, $\lambda_a \in \mathbb{R}^+$, and are nonzero only for samples on the decision boundary. The full theorem and requirements are provided in \cref{app:kkt}. 
We can apply \cref{eq:kkt_conv} to the quadratic network as 
\begin{align}
    \!\!\!\!
    \tilde{v} = \sum_{a=1}^N \lambda_a y_a (\tilde{W} x_a)^2,~
    \tilde{W} = \sum_{a=1}^N \lambda_a y_a \tilde{v}^Tx_a (W x_a),
\end{align}
where we assume no biases (consistent with the case of similar spectrum and different basis). 
We posit, that in high dimensions $\lambda_a \propto N^{-1}$, since at large $d$ every sample in the dataset should lie on the margin and contribute an equal amount, and the decision hyper-surface area should be comparable to the entire volume~\citep{hsu2022proliferation}. Under this ansatz, the above equations can be solved numerically, and indeed approximate the BOC as $d_h$ increases.
We present evidence that the Bayes optimal decision boundary can be realized by a two-layer neural network with quadratic activation, and increasing the number of hidden units increases the probability of converging to this minima under gradient flow in~\cref{fig:network_optimal}.

\vspace{-.2cm}
\section{Results Extending to Realistic Data and Networks}
\label{sec:Realistic_optimal}
\vspace{-.2cm}

\begin{figure*}[t!]
    \centering
    \includegraphics[width=0.965\linewidth]{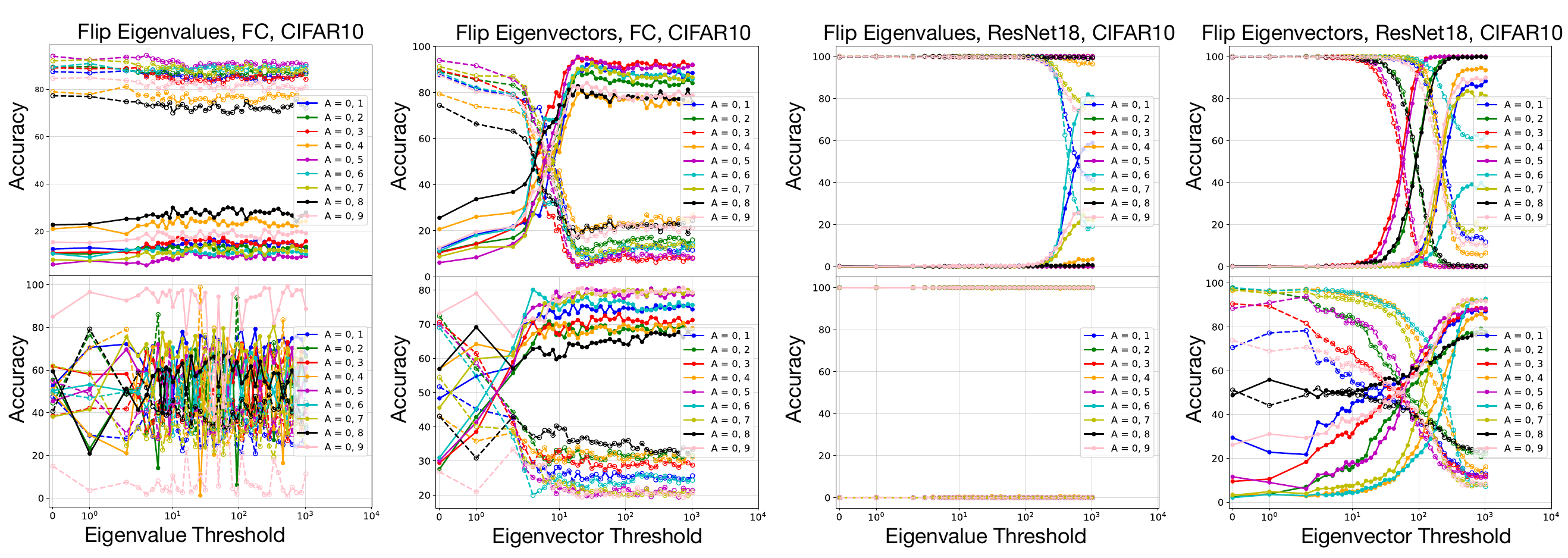}
    \includegraphics[width=.965\linewidth]{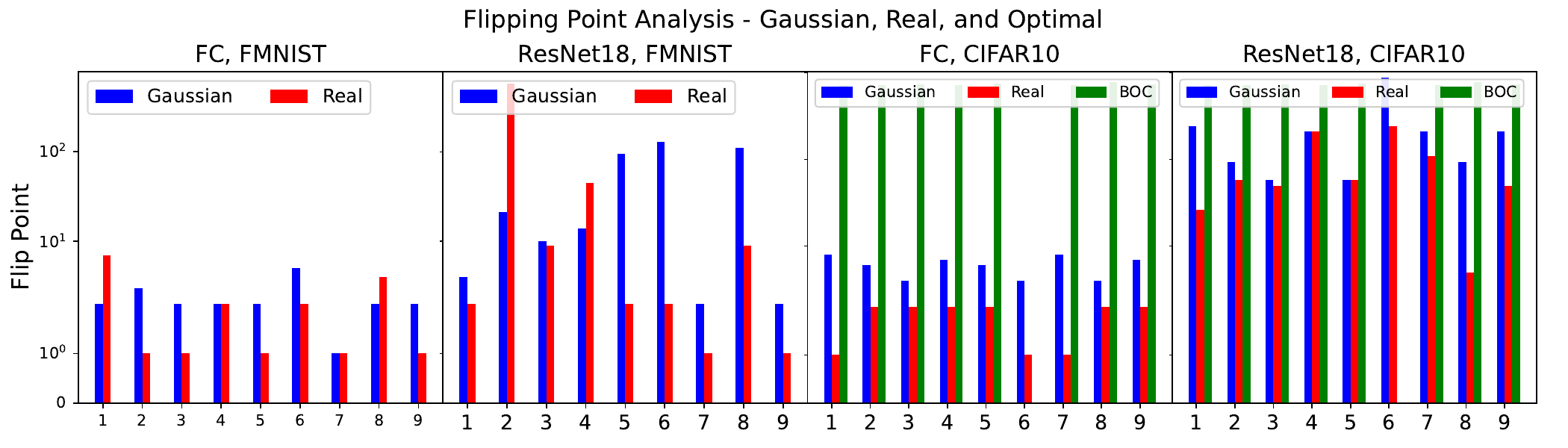}
        \vspace{-.4cm}
    \caption{The classification flipping test on the FMNIST and CIFAR10 datasets between class 0 and other classes. {\bf Top row:} training and tests on GMMs generated from the class covariance matrices. 
     {\bf Middle row:} training and tests on real images after whitening, rescaling and coloring. 
     {\bf Bottom row:} eigenvector threshold at the flipping point, for FMNIST and CIFAR10, using GMMs and real images, when either training with a NN or predicting with the BOC. {\bf Blue} columns indicate the flipping point on GMM classification, {\bf red} indicates the same test on real images, and {\bf green} shows the predictions of the BOC given only by information on the covariance matrices, as given by~\cref{eq:beta_def}. 
     The BOC is only shown for CIFAR10, due to numerical instabilities.
    }
    \vspace{-0.5cm}
\label{flipping_tests}
\end{figure*}

In the previous section, we showed that small differences in eigenvalues and eigenvectors are magnified by the dimension, making classification easier. 
We further demonstrated that differences in the eigenvectors are faster growing in the dimension of the data, suggesting eigenvectors play a more significant role in the Bayesian decision boundary. Next, we will provide clear evidence that these conclusions hold, and extend to real-world datasets, by performing a set of tests related to the various properties of the covariance matrices. 
We train two common network architectures to perform binary classification, both on GMMs and on real images, exploring their performance as we change the covariance structure of each class.

Concretely, we consider two popular network architectures: fully connected (FC) and convolutional neural networks (CNN), tested on the CIFAR10~\citep{cifar10} and Fashion-MNIST~\citep{xiao2017} datasets. For each training procedure, the model is optimized to classify samples from two classes in the given dataset (class 0 versus classes 1-9 aggregated).

The optimization procedure proceeds as follows: for each class, the samples are split into training and evaluation subsets. We then compute the covariance matrix of the training and evaluation subsets separately for the first and second classes. New synthetic data is generated by sampling from a multivariate Gaussian distribution with zero mean and the corresponding covariance matrix. This synthetic data is used to train the model to distinguish between the two classes (i.e. classify the Gaussians). When trained on real images, the model is tested on real images and not gaussians.

The optimization objective is the binary cross-entropy loss. The FC architecture consists of 3 dense layers with 2048 units each utilizing ReLU activations, followed by a softmax output layer. For CNNs, we employ ResNet-18~\citep{he2015deep}. All models are optimized with the Adam optimizer, a learning rate of 0.001, and Cosine Annealing LR scheduling over 50 epochs with a batch size of 256. The synthetic datasets contain 1,280 images per class, split 80-20 for training and validation. Our computational resource was a single NVIDIA GeForce GTX 1650 GPU, with 16GB RAM.

\vspace{-.2cm}
\subsection{$\Delta \alpha$ Tests on GMMs}
\vspace{-.2cm}

We first perform an experiment to quantitatively evaluate a network's ability to utilize covariance matrix eigenvalue scaling for classification tasks. Specifically, two random covariance matrices are constructed using~\cref{CM}, with one covariance has the spectral scaling of $\alpha$ and the other of $\alpha + \Delta\alpha$. Critically, both matrices share the same basis of eigenvectors. Approximately 500 samples are then drawn from each covariance matrix.

A model is trained to classify between the two classes defined by the distinct covariance matrices. ~\cref{fig:model_FC_alpha_dalpha} shows the results for a fully connected network, as well as for a ResNet18 architecture~\citep{he2015deep}. To assess the performance variability, the results are averaged over an ensemble of five independent training runs.

By varying only the scaling exponent between the covariances while keeping the eigenvectors fixed, this experiment protocol directly tests a network's capability to leverage the eigenvalue scaling imparted by the covariance matrix for discriminative learning. The stability of the classifications indicate the degree to which these deep models can extract and utilize the intrinsic geometrical information encoded in the covariance descriptors.

\vspace{-.2cm}
\subsection{Flip Tests on GMMs Constructed from Real Data}
\vspace{-.2cm}

Next, we consider the role of eigenvectors and eigenvalues in classifying overlapping GMMs, as well as real images.
It is known that the eigenvectors and eigenvalues of the data covariance matrix reveal different aspects of the data. 
In order to understand which of the two contains pertinent information for NN classification performance, we consider the following setup:
we study two datasets, with covariance matrices $C_1$ and $C_2$, and decompose each one of them to a set of eigenvectors and eigenvalues: $\left\{ \lambda^{1}_i, v^{1}_i \right\}$, $\left\{ \lambda^{2}_i, v^{2}_i \right\}$.
We then construct a new covariance matrix by combining some of the eigenvalues and eigenvectors of $C_1$ and $C_2$. To quantify the ratio of eigenvectors/eigenvalues coming from each class, we define two threshold indices $\tau_\lambda$ for the eigenvalues and $\tau_v$ for the eigenvectors , where each of them is an integer between 0 and the dimension of $C_1,C_2$, given by $d$. The new covariance matrix reads:
\begin{gather}
    C_{\tau_\lambda, \tau_v} = V^T \Lambda V \ ,
\end{gather}
where $V = V_{\tau_v}$ and $\Lambda = \Lambda_{\tau_\lambda}$ are the new basis, and eigenvalues, respectively, which are composed by a mixture between the two classes:
\begin{gather}
    v_i = \begin{cases}
        {v^{1}}_i & i \leq \tau \\
        {v^{2}}_i & \tau < i \\
    \end{cases},
    \hspace{1cm}
    \lambda_i = \begin{cases}
        {\lambda^{1}}_i & i \leq \tau \\
        {\lambda^{2}}_i & \tau < i  \ .
    \end{cases}
\end{gather}
This process is denoted as a {\it flip test}, illustrated in~\cref{fig:FlipTests}.

It is worth noting that when composing the rotation matrix $V$, we must ensure that it is nearly orthogonal to maintain the properties of a true basis, which is not guaranteed in our scheme. However, we find that this matrix is very close to orthogonal in the following sense: we define the following error, and require $e\ll1$
\begin{align}
\label{eq:error}
    e \equiv \max_\tau{ \tfrac{1}{d} \left||V^\dag V - I |\right|_F },
\end{align}
where $\|\cdot\|_F$ is the Frobenius norm.
\cref{eq:error} sets an upper bound for the error over all of the thresholds $\tau \in [0, d]$. We practically found that $e \sim 0.018$, while computing it for a randomly generated matrix composed of unit norm vectors gives $e\sim 1$.

In \cref{flipping_tests} (top row), we show the results of training a FC and ResNet18 networks to classify GMMs constructed from different $C_1,C_2$ and tested on samples drawn from $C_{\tau_\lambda, \tau_v}$. The baseline results are that the network classifies samples as coming from $C_1$ if $C_{\tau_\lambda, \tau_v}$ has both the eigenvalues and eigenvectors of $C_1$. When constructing $C_{\tau_\lambda, \tau_v}$ with the eigenvectors of $C_1$ and the eigenvalues of $C_2$, the network still classifies samples as coming from $C_1$. This is not the case when taking the eigenvalues from $C_1$ but some of the eigenvectors from $C_2$. Evidently, it requires a $\mathcal{O}(100)$ eigenvectors from $C_2$ to convince the network that the samples came from $C_2$. 

In \cref{flipping_tests} (bottom row), we show that this result is architecture dependent, but the existence of an eigenvector flipping point persists throughout our experiments. We further compare this result with the BOC prediction, showing that indeed the BOC requires more eigenvectors to be flipped before determining that samples belong in $C_2$ rather than $C_1$. This is shown in detail in \cref{fig:flip_point_boc}.

We further analyze the above flip tests for different data sizes used to construct $C_1$ and $C_2$. We find that the results do not change above a certain number of samples, as the covariance matrices are well approximated beyond that point. \cref{fig:c1_vs_c2_accuracy_vs_nos_CIFAR10} shows the results.






\vspace{-.2cm}
\subsection{Flip Tests on Real Data}
\vspace{-.2cm}

To ensure that the overlapping mixture model holds for real-data, we examined the effect of adding the vector class means computed from the CIFAR10 and FMNIST datasets. We 
observed no qualitative change in the results, indicating that the assumption of zero mean is 
justified in these cases.

In order to perform the same flipping tests on real images, we simply perform a whitening transformation, followed by a rescaling, and then rotating to the basis of choice~\citep{eigenfaces97}. For an illustration of the resulting images see ~\cref{fig:rotated_images_comparison_ResNet18_FMNIST} and \cref{fig:rotated_images_comparison_ResNet18_CIFAR10}.
This process ensures that the class covariance matrices of the data have been matched to our design, but inherently affects all the higher moments of the underlying distribution in an unpredictable way. Interestingly, as can be seen in~\cref{flipping_tests} (top row), the occurrence of a clear flipping point which depends on eigenvectors and not eigenvalues is manifest in real images. In~\cref{flipping_tests} (bottom row) we show a comparison between the thresholds for GMMs and real images, where the results indicate a close similarity, particularly for ResNet18 trained on CIFAR10, which can be attributed to its high expressivity.


\vspace{-.3cm}
\section{Conclusions and Limitations}
\label{conclusions}
\vspace{-.2cm}

We studied binary classification of high dimensional overlapping GMM data
as a framework to model real-world image datasets and to quantify the importance for the classification tasks of
the eigenvalues and the eigenvectors of the data covariance matrix. 
We showed that deep neural networks trained for classification, learn predictors that approximate the Bayes optimal classifiers, and 
demonstrated that the decision thresholds for networks trained on authentic data correlate with the covariance eigenvectors rather than the eigenvalues, compatible with the GMM analysis. Our results
reveal new theoretical insights to the neural networks' ability to perform probabilistic inference and distill statistical patterns from complex distributions.

{\bf{Limitations:}} 
Firstly, we focused on the $\gamma \ll1$ regime of the estimated class covariances, while in many real-world cases, the empirical limit may be more applicable~\cite{levi2023universal}.
In~\cref{sec:kkt_convergence_net}, our ansatz neglected the dependence of $\lambda$ on $d/N$ and the spectral density, which nonlinearly determines the number of points which lie on the decision surface. It might be possible to derive a similar result to~\cite{hsu2022proliferation} by taking a feature map approach to the quadratic net, but we postpone this to future work.
Finally, while we quantified the significance of the detailed structure of the data covariance matrix, i.e. its eigenvalues and eigenvectors and their relative importance for classification, we are still left with the open question regarding the
relative importance of the higher moments of the distribution, which we leave for a future study.

\section{Acknowledgements}
\label{ack}

We would like to thank Amir Globerson, Yohai Bar-Sinai, Yue Lu and Bruno Loureiro for useful discussions.
NL would like to thank G-Research for the award of a research grant, as well as the CERN-TH department for their hospitality during various stages of this work.
The work of Y.O. is supported in part by Israel Science Foundation Center of Excellence.



\bibliography{ML_Cov}

\bibliographystyle{unsrtnat}

\newpage
\onecolumn

\appendix

\setlength{\belowdisplayskip}{6pt} 
\setlength{\belowdisplayshortskip}{6pt}
\setlength{\abovedisplayskip}{6pt} 
\setlength{\abovedisplayshortskip}{6pt}

\section{Neural Collapse on Overlapping GMM Data}
\label{app:NC}

\begin{figure}[h!]
    \centering
    \includegraphics[width=.5\linewidth]{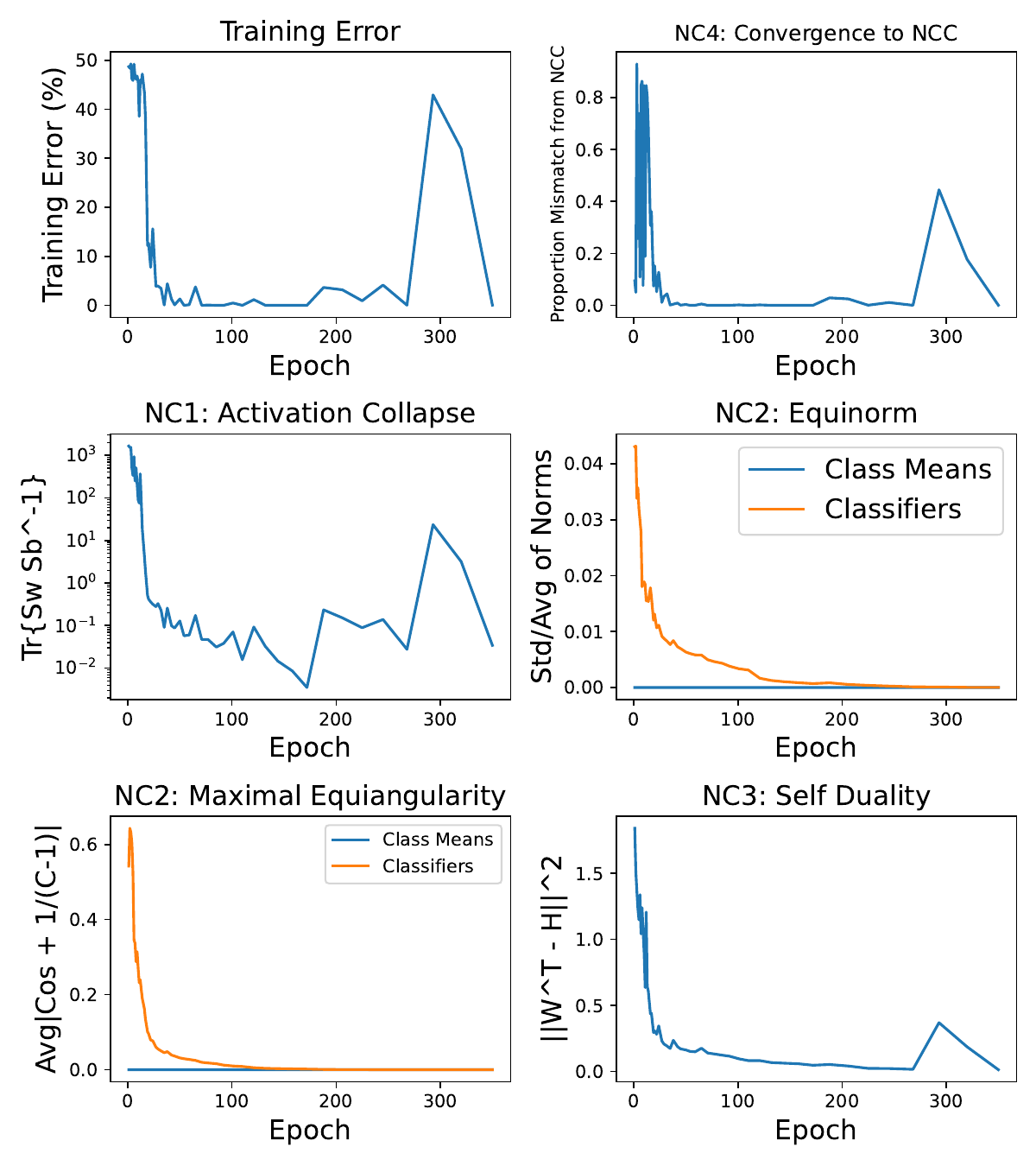}~
    \includegraphics[width=.5\linewidth]{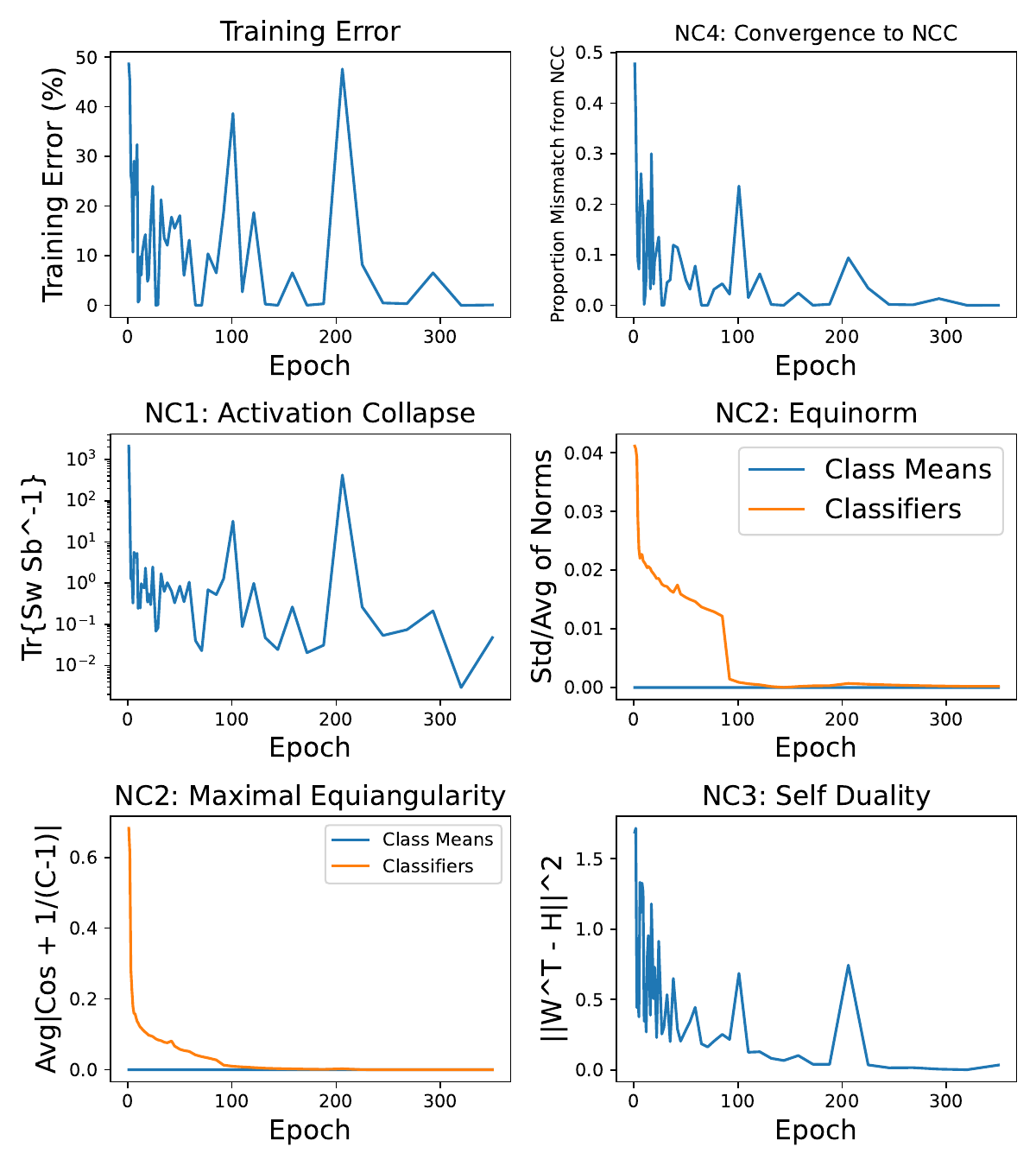}
    \caption{Results for neural collapse convergence for training ResNet18 on two classes (0,7) sampled from Gaussian versions of CIFAR10 data.
    Left: setting $\mu_A=\mu_B=0$. Right: $\mu_A\neq \mu_B$, as given by the class means.
    The network attains 99.9\% test accuracy in both cases.
    We see no substantial difference in convergence to neural collapse metrics, affirming our claims in the main text.}
    \label{fig:neural_collapse_no_means}
\end{figure}

Here, we explore the concept of Neural Collapse (NC) on overlapping GMMs generated from real data. Practically, we employ the metrics given in~\citet{Papyan_2020} to measure the NC when training a ResNet18 on gaussian versions of CIFAR10 images. In \cref{fig:neural_collapse_no_means}, we show that NC occurs on overlapping GMMs even when the class mean vector of the data has been set to zero, indicating that expressive networks are able to perform nonlinear transformations that generate class means, where none were present, and at the final layer still perform linear classification.

Various metrics of Neural Collapse, as given by~\citet{Papyan_2020}, are shown for GMMs generated from two classes of CIFAR10.
Here, we use 5000 samples per class, training a ResNet18 to perform binary classification using cross-entropy loss, on GMMs generated from greyscale versions of CIFAR10 images.
Each sample has dimension $1024$, and the training is done using the default specifications provided in the supplementary code of~\citet{Papyan_2020}

\section{KKT formulation}
\label{app:kkt}

As discussed in the main text, we posit that the approach towards a BOC for a NN can be explained by convergence to a KKT point.

Our reasoning follows \thmref{thm:known KKT} below, 
which holds for \emph{gradient flow} (i.e., gradient descent with an infinitesimally small step size). 
Before stating the theorem, we need the following definitions:
(1) We say that 
gradient flow
{\em converges in direction} to $\tilde{\btheta}$ if $\lim_{t \to \infty}\frac{\btheta(t)}{\norm{\btheta(t)}} = \frac{\tilde{\btheta}}{\norm{\tilde{\btheta}}}$, where $\btheta(t)$ is the parameter vector at time $t$;
(2) We say that a network $\Phi$ is \emph{homogeneous} w.r.t. the parameters $\btheta$ if there exists $L>0$ such that for every $\lambda>0$ and $\btheta,\bx$ we have $\Phi(\alpha \btheta; \bx) = \lambda^L \Phi(\btheta; \bx)$. 
Thus, scaling the parameters by any factor $\lambda>0$ scales the outputs by $\lambda^L$. 
We note that essentially any fully-connected or convolutional neural network with ReLU activations is homogeneous w.r.t. the parameters $\btheta$ if it does not have any skip-connections (i.e., residual connections) or bias terms, except possibly for the first layer.

\begin{theorem}[Paraphrased from \cite{lyu2020gradient,ji2020directional}]
\label{thm:known KKT}
	Let $\Phi(\btheta;\cdot)$ be a homogeneous 
	ReLU neural network. Consider minimizing 
	the logistic loss over a binary classification dataset $ \{(\bx_a,y_a)\}_{a=1}^N$ using gradient flow. Assume that there exists time $t_0$ such that $\cl(\btheta(t_0))<1$\footnote{This ensures that $\ell(y_a \Phi(\btheta(t_0); \bx_a)) <1$ for all $i$, i.e. at some time $\Phi$ classifies every sample correctly.}.
	Then, gradient flow converges in direction to a first order stationary point (KKT point) of the following maximum-margin problem:
\begin{equation}
\label{eq:optimization problem}
	\min_{\btheta'} \frac{1}{2} \norm{\btheta'}^2 \;\;\;\; \text{s.t. } \;\;\; \forall i \in [n] \;\; y_a \Phi(\btheta'; \bx_a) \geq 1~.
\end{equation}
Moreover, $\cl(\btheta(t)) \to 0$ 
as $t \to \infty$.
\end{theorem}

The above theorem guarantees directional convergence to a first order stationary point (of the optimization problem ~(\cref{eq:optimization problem})), which is also called \emph{Karush–Kuhn–Tucker point}, or \emph{KKT point} for short. 
The KKT approach allows inequality constraints, and is a generalization of the method of \emph{Lagrange multipliers}, which allows only equality constraints.

The great virtue of Theorem \ref{thm:known KKT} is that it characterizes the \emph{implicit bias} of gradient flow with the logistic loss for homogeneous networks. Namely, even though there are many possible directions of $\frac{\btheta}{\norm{\btheta}}$ that classify the dataset correctly, gradient flow converges only to directions that are KKT points of Problem~(\cref{eq:optimization problem}). In particular, if the trajectory $\btheta(t)$  
of gradient flow under the regime of Theorem \ref{thm:known KKT} converges in direction to a KKT point $\tilde{\btheta}$, then 
we have the following:
There exist $\lambda_1,\ldots,\lambda_n\in\reals$ such that 
\begin{align} 
    &\tilde{\btheta} = \sum_{a=1}^N \lambda_a y_a \nabla_\btheta  \Phi(\tilde{\btheta}; \bx_a)~ &\text{(stationarity)}\label{eq:stationary}\\
    & \forall a \in [N], \;\; y_a \Phi(\tilde{\btheta}; \bx_a) \geq 1 &\text{(primal feasibility)}\label{eq:prim feas} \\
    &\lambda_1,\ldots,\lambda_N \geq 0 &\text{(dual feasibility)}\label{eq:dual feas}\\
    & \forall a \in [N],~~ \lambda_a= 0 ~ \text{if}~  y_a \Phi(\tilde{\btheta}; \bx_a) \neq 1 & \text{(complementary slackness)}\label{eq:comp slack}
\end{align}
Our main insight is based on \cref{eq:stationary}, which implies that the parameters $\tilde{\btheta}$ are a linear combinations of the derivatives of the network at the training data points. 
We say that a data point $\bx_a$ is \emph{on the margin} if $y_a \Phi(\tilde{\btheta}; \bx_a) = 1$ (i.e. $|\Phi(\tilde{\btheta}; \bx_a)|=1$) .
Note that \cref{eq:comp slack} implies that only samples which are on the margin 
affect \cref{eq:stationary},
since samples not on the margin have a coefficient $\lambda_a=0$. In sufficiently high dimension we expect no samples to be off the margin in GMM classification therefore all $\lambda_a$ should be nonzero and contribute a similar amount.


\section{Dataset Properties}
\label{app:data_properties}

\begin{figure*}[h!]
    \centering
    \includegraphics[width=1 \linewidth]{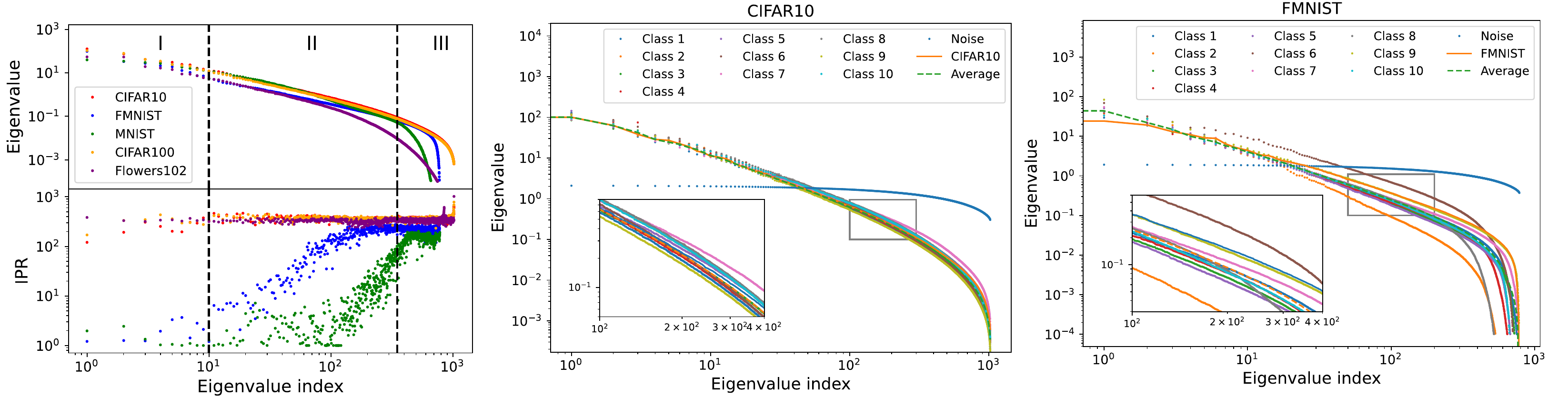}    
        \vspace{-.4cm}
    \caption{{\bf Left:} Partition the eigenvector indices to three regimes: the large eigenvalues, the bulk and the small eigenvalues. The top figure shows the eigenvalues versus their index, and the bottom show the corresponding IPR versus the eigenvalues index.
    {\bf Center/Right:} The eigenvalues of the covariance matrix for the different classes in CIFAR10/FMNIST dataset, with a comparison to uncorrelated data ({\bf blue}). The {\bf orange} line represents the eigenvalues of the covariance matrix of the whole dataset and the {\bf green} dashed line shows the average of the different classes. We see the three regimes of eigenvalues: the large eigenvalues - up to  $\pm 10$ first eigenvalues, the bulk scaling - between $\pm 10$ to $\pm 300$, and the small eigenvalues - from $\pm 350$.}
    \label{fig:eigenvalues_CIFAR10}
\end{figure*}

We show the eigenvalues scaling structure and the 
Inverse Participation Ratio (IPR) of the eigenvectors.
We separate the eigenvalues to three parts: I. The large eigenvalues, II. The bulk eigenvalues, and III. The small eigenvalues, as can be seen in \cref{fig:eigenvalues_CIFAR10}.
These three regimes are seen also in real datasets as shown in previous work \citep{levi2023underlying,levi2023universal}. \cref{fig:eigenvalues_CIFAR10} shows the eigenvalues for each of the classes in the FMNIST and CIFAR10 datasets, respectively.

\begin{figure}[h!]
    \centering
    \includegraphics[width=.5\linewidth]{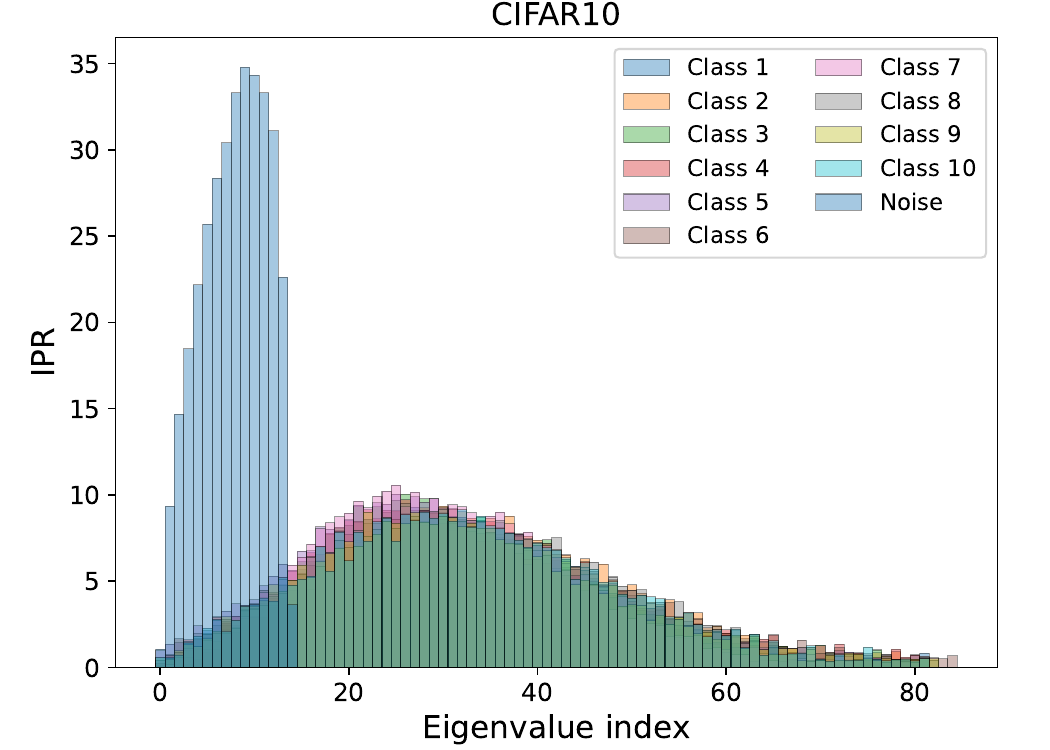}~
    \includegraphics[width=.5\linewidth]{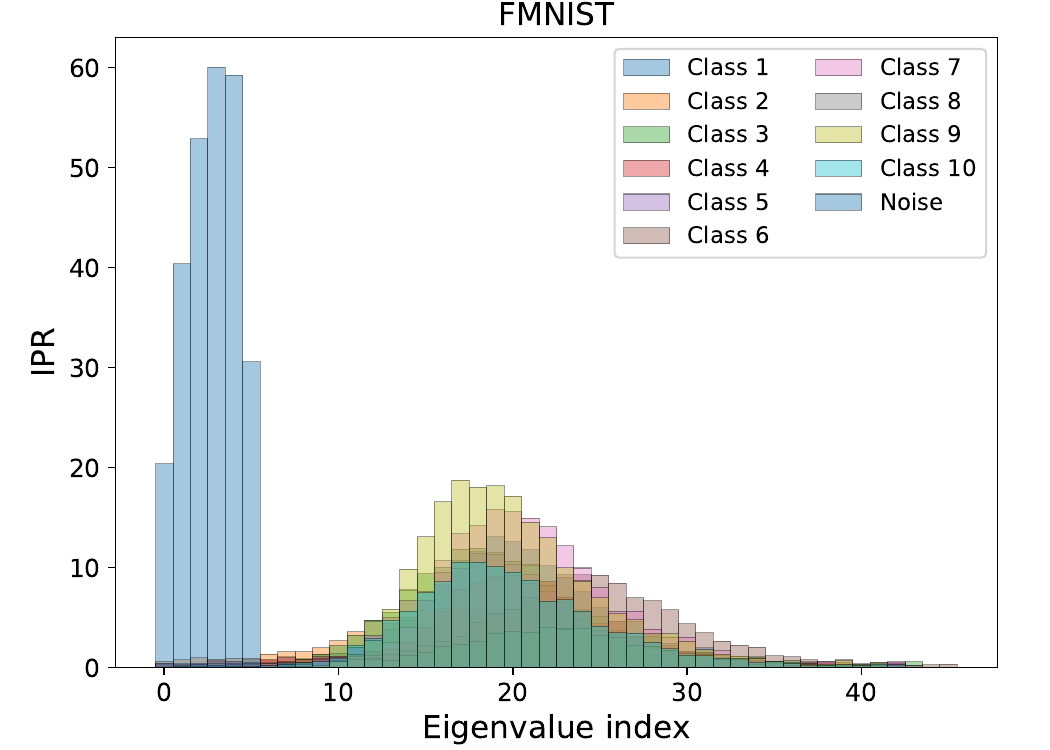}
   \caption{The histogram of the IPR of the covariance matrix eigenvectors for the different classes in the CIFAR10 dataset (left), and FMNIST (right), with a comparison to noise (blue). }
    \label{fig:IPR_distribution_CIFAR10}
\end{figure}

The IPR defined as:
\begin{align}
    IPR \equiv \left( \sum_n \left|v_n\right|^4 \right)^{-1} ,
\end{align}
is used to measure the localization rate (entropy) of a vector.
\cref{fig:IPR_distribution_CIFAR10} shows the 
IPR of the eigenvectors of the covariance matrices for 
the different classes in the CIFAR10 dataset and FMNIST.
As can be seen from the Figure, the changes in the eigenvalues structure and the IPR is small.

\section{Appendix - Flipping tests examples}
\label{app:flipping_test}

\begin{figure}[htbp!]
    \centering
    \includegraphics[width=1\linewidth]{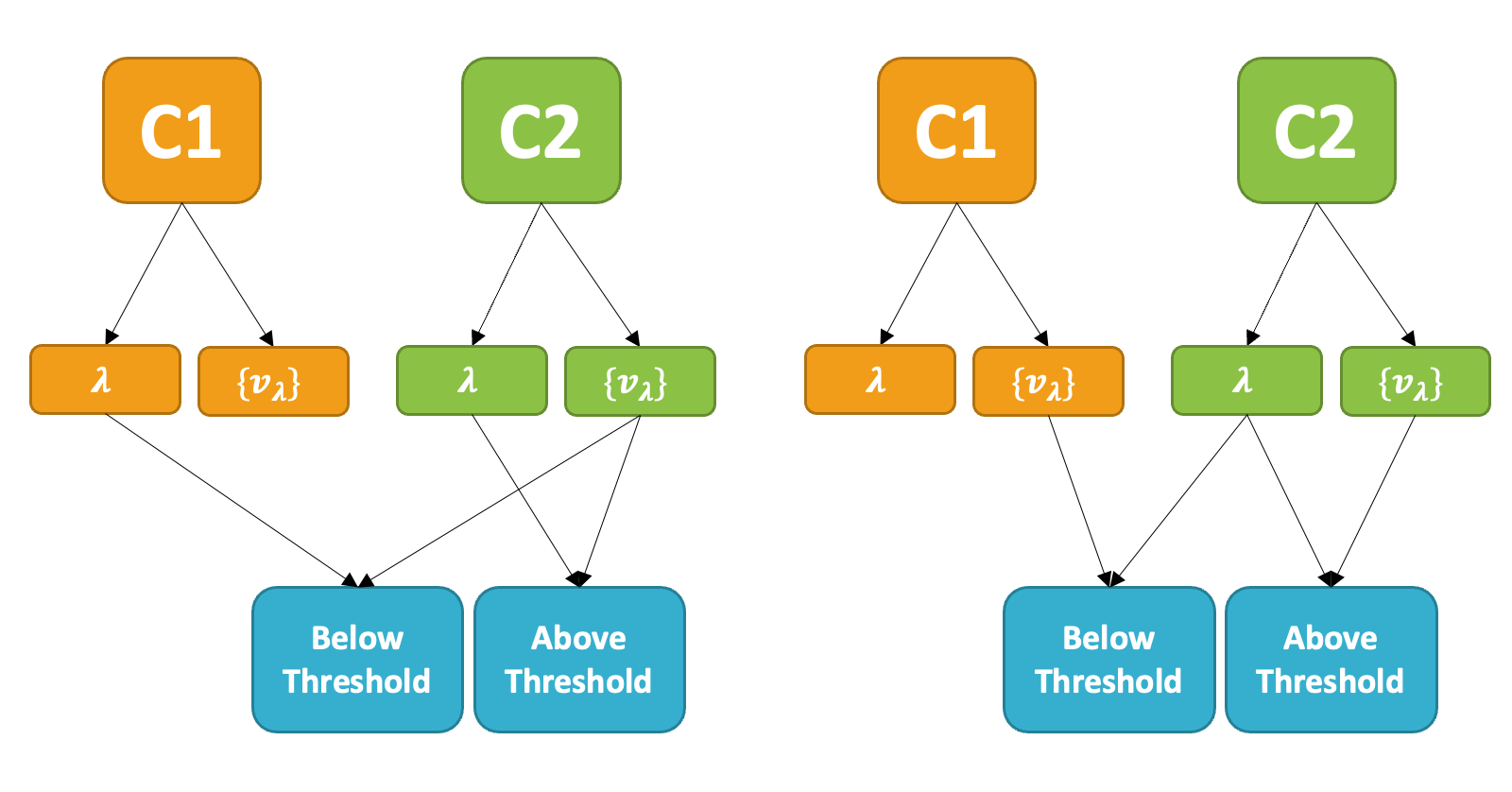}
    \caption{The classification flipping tests. We evaluate the model classification performance on different covariance matrices. Left - building a new covariance matrix by flipping the eigenvectors up to a certain threshold. Right - building a new covariance matrix by flipping the eigenvalues up to a certain threshold.  This method provides information about the relative importance of the different eigenvectors and eigenvalues for the model to make a classification decision.}
    \label{fig:FlipTests}
\end{figure}

\begin{figure}[h!]
    \centering
    \includegraphics[width=1\linewidth]{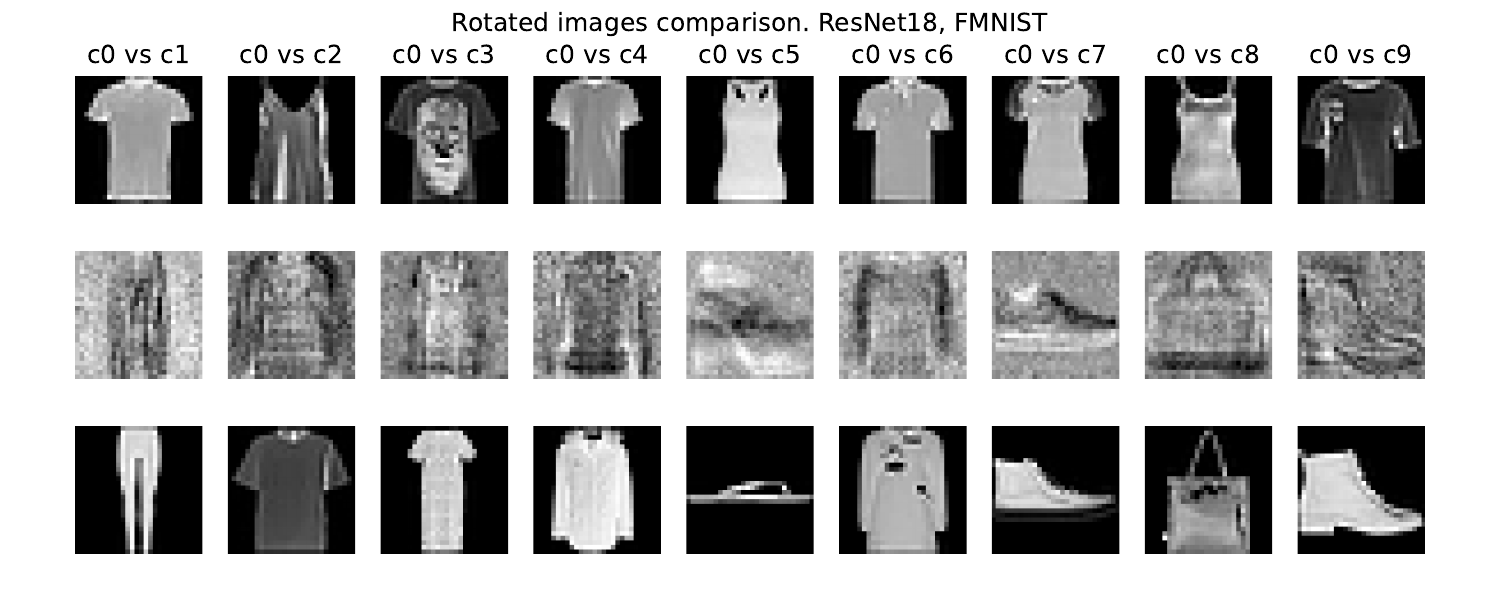}
    \caption{Examples of rotated images in the FMNIST dataset. Each of the samples of class 0 (shirt, top row), was rotated using the covariance matrix of another class (bottom row), and the middle row shows the result of the rotation. In all of our tests, for any rotation of class 0 sample, with class $c$ basis, the model classified the rotated image as a sample from class $c$.}
    \label{fig:rotated_images_comparison_ResNet18_FMNIST}
\end{figure}

\begin{figure}
    \centering
    \includegraphics[width=1\linewidth]{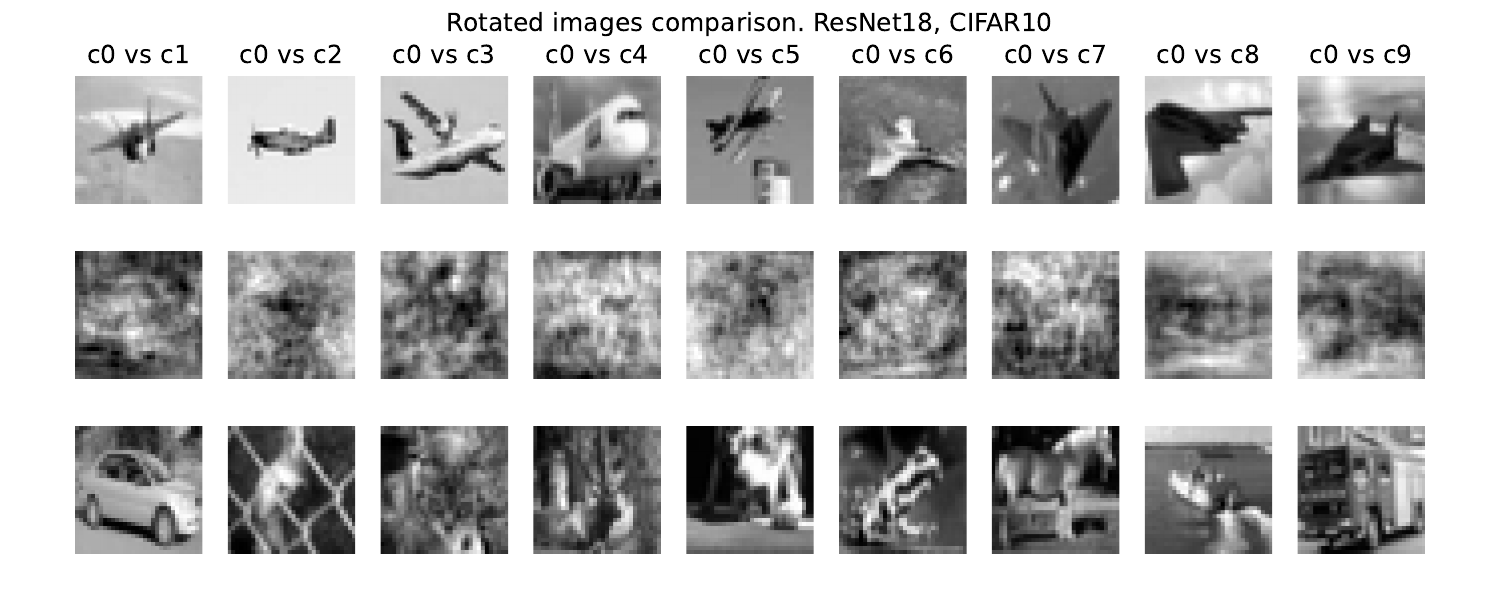}
    \caption{Examples of rotated images in the CIFAR dataset. Each of the samples of class 0 (airplane, top row), was rotated using the covariance matrix of another class (bottom row), and the middle row shows the result of the rotation. In all of our tests, for any rotation of class 0 sample, with class $c$ basis, the model classified the rotated image as a sample from class $c$.}
    \label{fig:rotated_images_comparison_ResNet18_CIFAR10}
\end{figure}

\begin{figure*}[h!]
    \centering  
    \includegraphics[width=.8\linewidth]{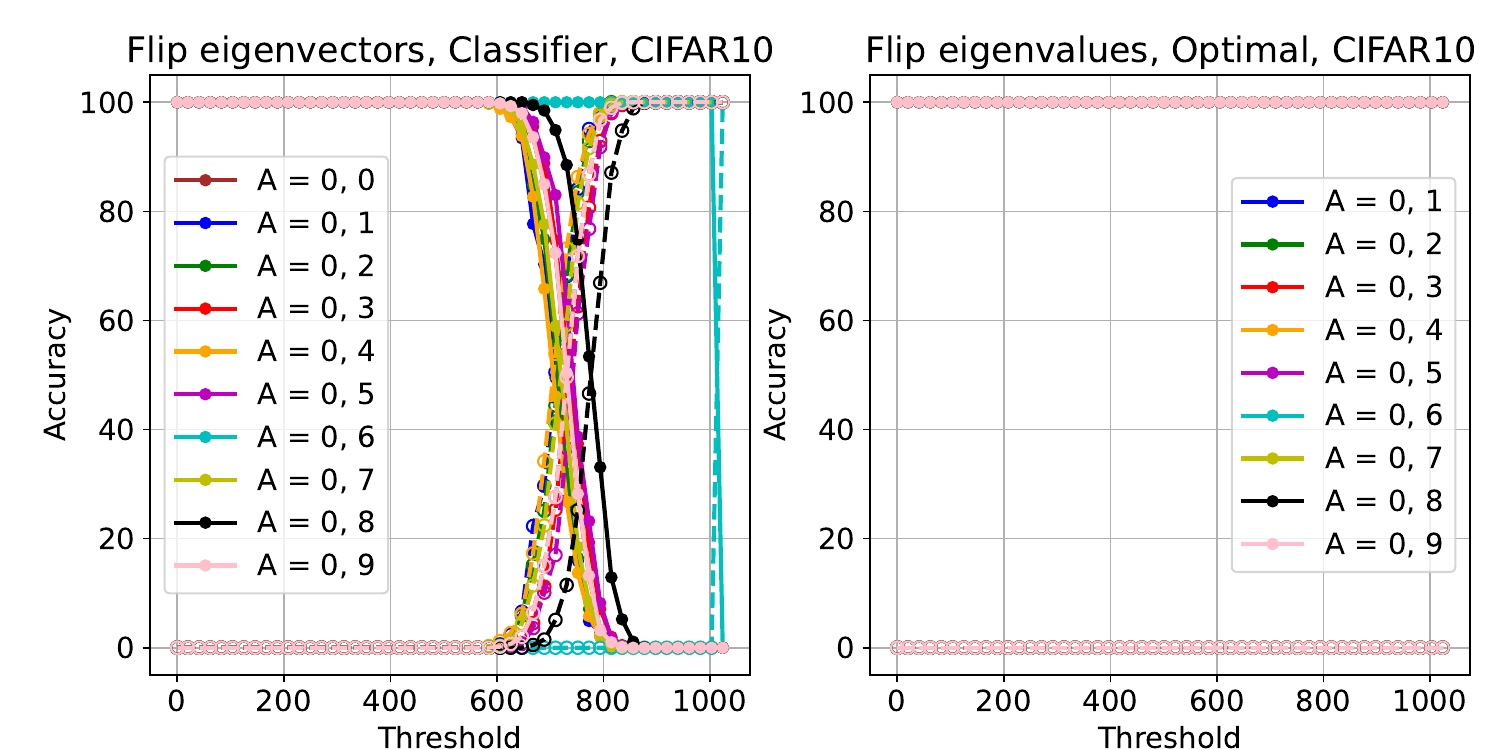}~
    \caption{The classification eigenvector (left) and eigenvalues (right) flipping test on the optimal classifier. CIFAR10 dataset, comparing class 0 with other classes.}
    \label{fig:flip_point_boc}
\end{figure*}

\begin{figure}[t!]
    \centering
    \includegraphics[width=0.5\linewidth]{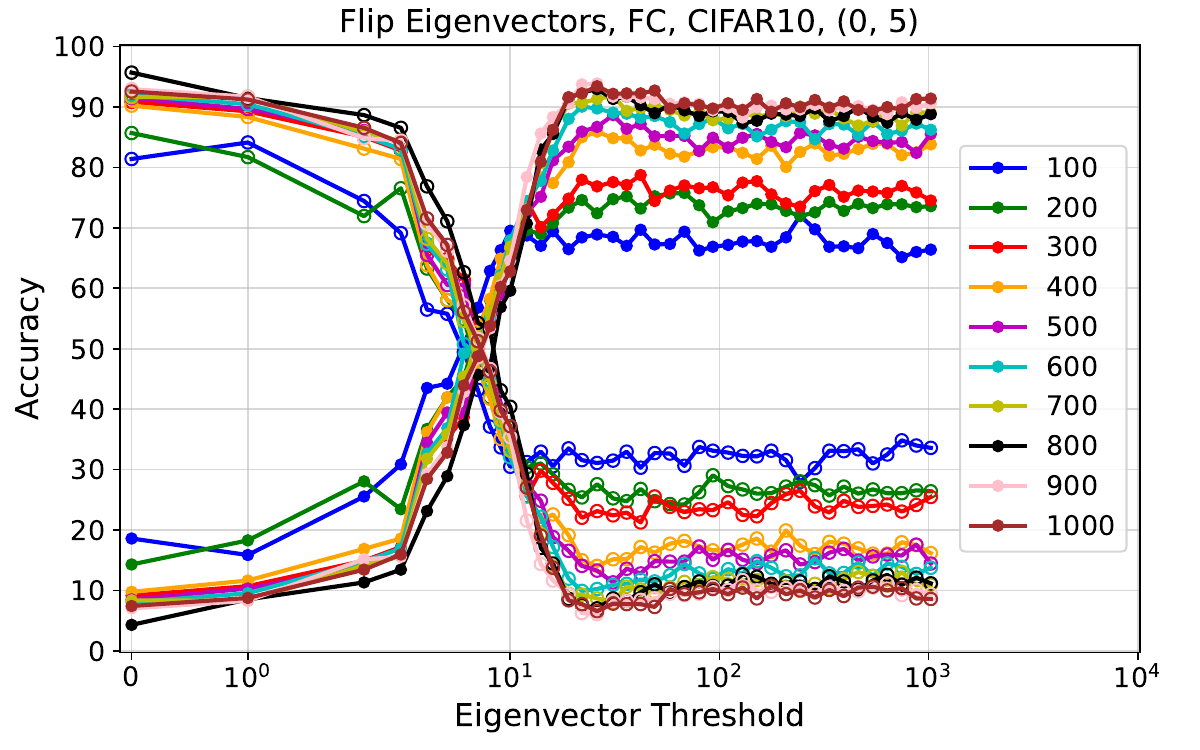}~
    \includegraphics[width=0.45\linewidth]{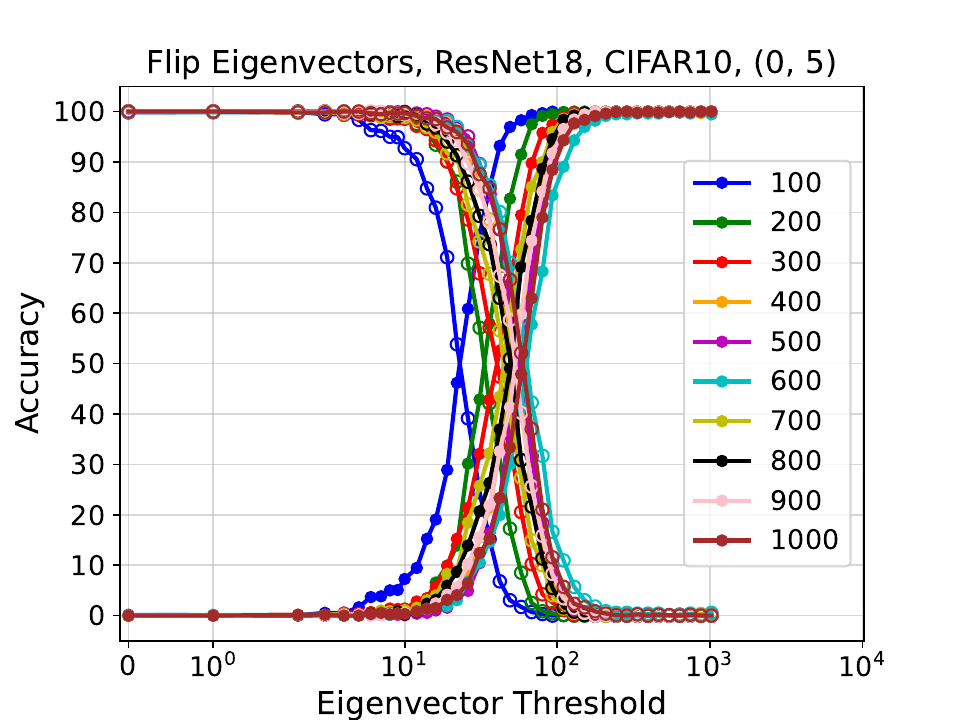}
    \caption{The classification eigenvector flipping test on the Gaussian dataset between class 0 and class 5, for different number of samples. FC model (left) and ResNet18 (right), CIFAR10 dataset.}
    \label{fig:c1_vs_c2_accuracy_vs_nos_CIFAR10}
\end{figure}

\begin{figure*}[t!]
    \centering
    \includegraphics[width=.45\linewidth]{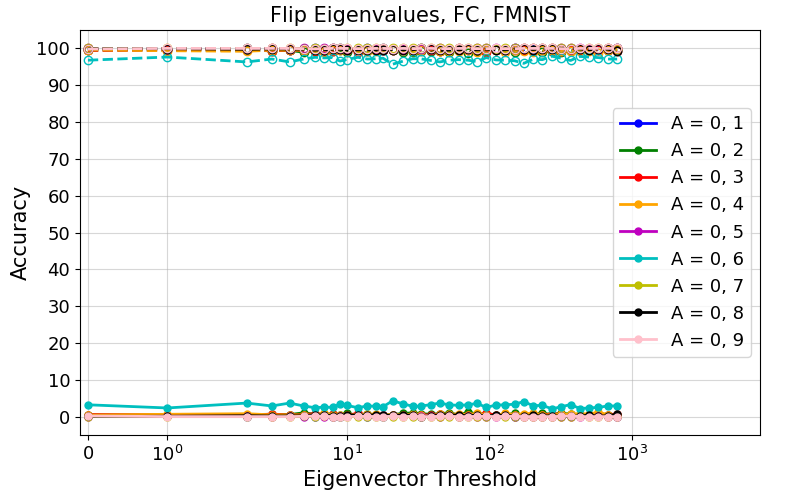}~
    \includegraphics[width=.45\linewidth]{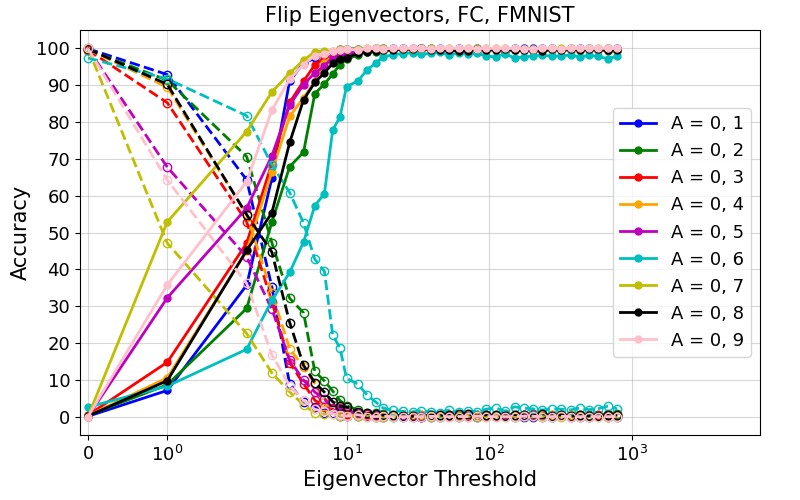}
    \includegraphics[width=.45\linewidth]{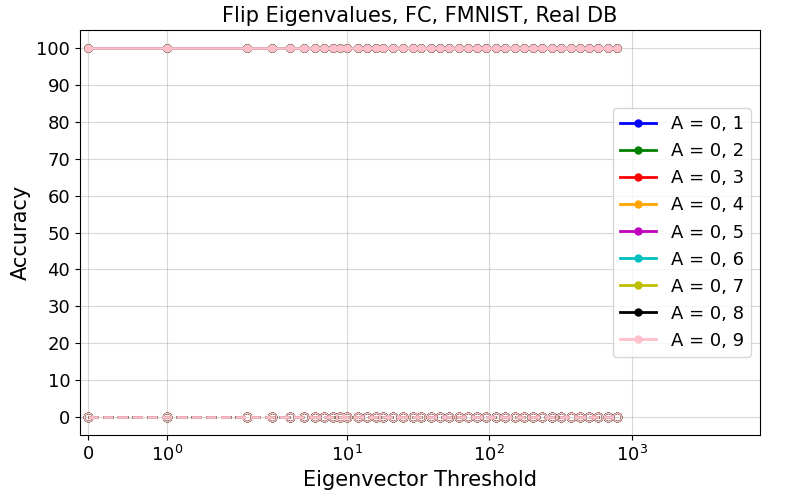}~
    \includegraphics[width=.45\linewidth]{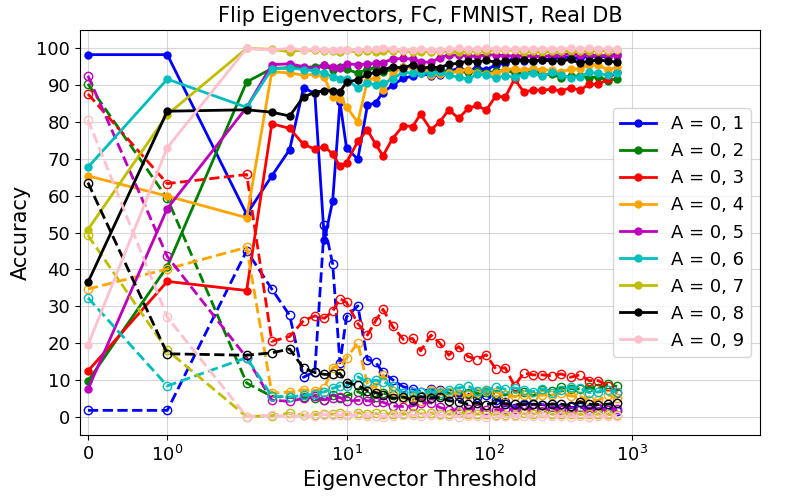}
    \caption{The classification eigenvalues flipping test on the rotated FMNIST dataset between class 0 and other classes for the FC model.}
\label{accuracy_vs_threshold_FMNIST_FalseTrue_rotateeig_fc}
\end{figure*}

\begin{figure*}[t!]
    \centering
    \includegraphics[width=.45\linewidth]{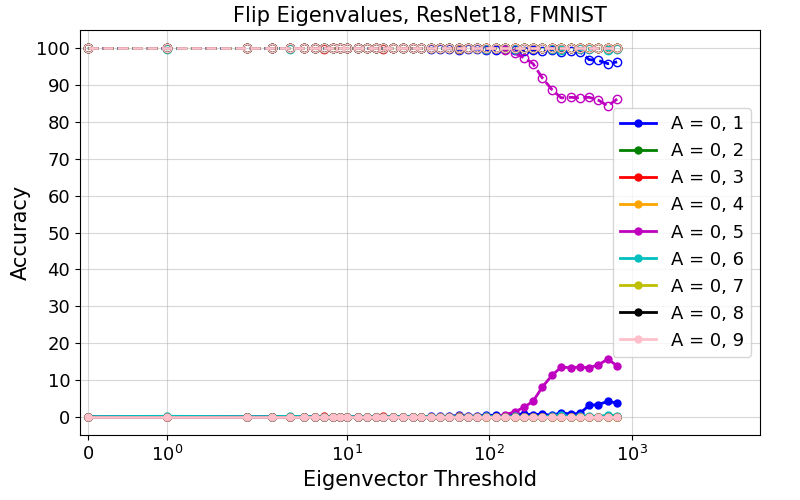}
    \includegraphics[width=.45\linewidth]{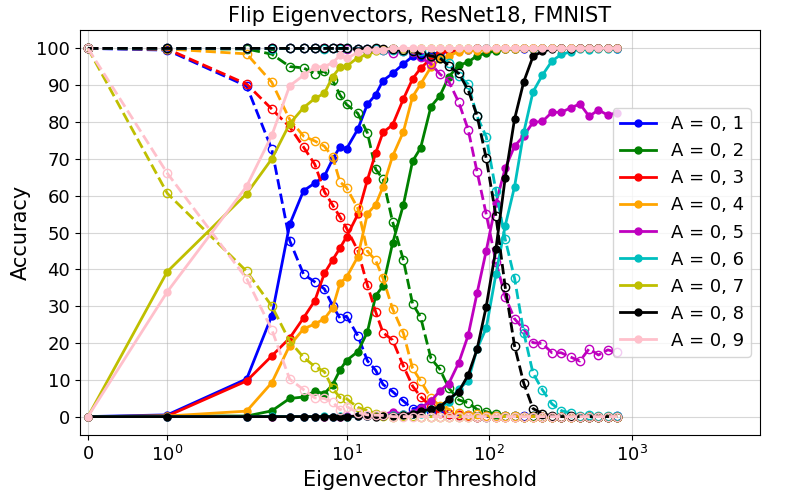}
    \includegraphics[width=.45\linewidth]{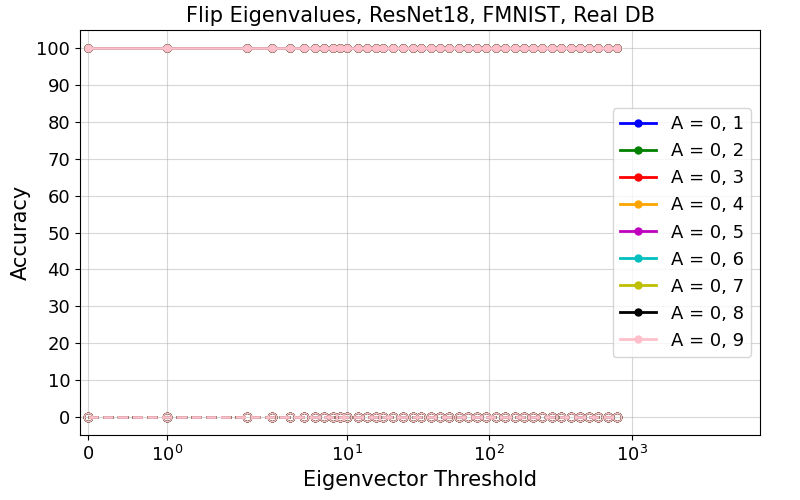}
    \includegraphics[width=.45\linewidth]{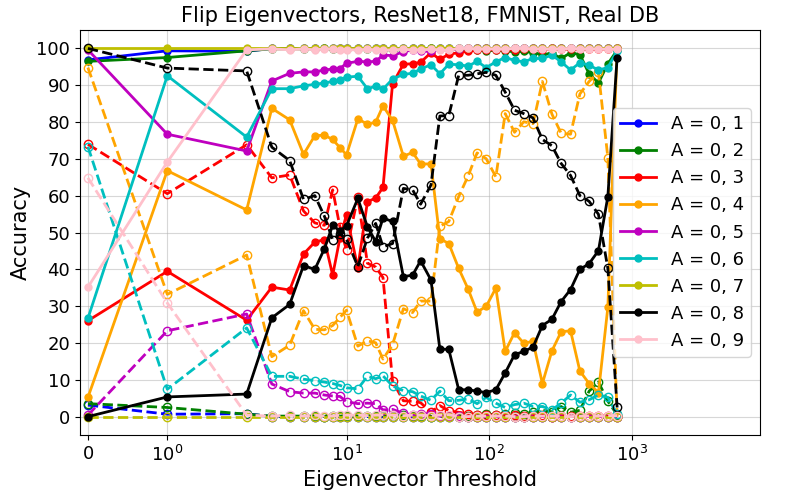}
    \caption{The classification eigenvalues flipping test on the rotated FMNIST dataset between class 0 and other classes for the ResNet 18 model.}
\label{accuracy_vs_threshold_FMNIST_FalseTrue_rotateeig_res}
\end{figure*}

We include the different figures concerning the classification flipping tests referred to in the main text.
The diagram in \cref{fig:FlipTests} outlines the structure of the flipping tests.
In \cref{fig:rotated_images_comparison_ResNet18_FMNIST} and
\cref {fig:rotated_images_comparison_ResNet18_CIFAR10} we see rotated images of FMNIST and CIFAR10, respectively.
In \cref{fig:flip_point_boc}, 
\cref{fig:c1_vs_c2_accuracy_vs_nos_CIFAR10} and 
\cref{accuracy_vs_threshold_FMNIST_FalseTrue_rotateeig_fc,accuracy_vs_threshold_FMNIST_FalseTrue_rotateeig_res} we plot the eigenvalues and eigenvectors thresholds.

The code for our experiments is provded in \url{https://github.com/khencohen/FlippingsTests}.

\end{document}